\documentclass[10pt,logo]{loopwamreport}
\definecolor{loopwamteal}{HTML}{039EA7}
\definecolor{loopwampurple}{HTML}{6050D1}
\newcommand{\LoopWAMLogo}{%
  \raisebox{-0.22\height}{\includegraphics[height=18pt]{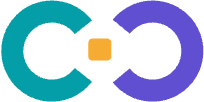}}%
  \hspace{5pt}%
  {\normalfont\bfseries\fontsize{17}{20}\selectfont
   \textcolor{loopwamteal}{Loop}\textcolor{loopwampurple}{WAM}}%
}
\usepackage[authoryear,sort&compress,round]{natbib}
\usepackage{parskip}

\usepackage{amsmath,amsfonts,bm}

\def\eqref#1{equation~\ref{#1}}

\def\1{\bm{1}}

\DeclareMathAlphabet{\mathsfit}{\encodingdefault}{\sfdefault}{m}{sl}
\SetMathAlphabet{\mathsfit}{bold}{\encodingdefault}{\sfdefault}{bx}{n}

\usepackage{graphicx}
\usepackage{booktabs}
\usepackage{url}
\usepackage{xcolor}
\usepackage{float}
\usepackage{pifont}
\usepackage{hyperref}
\newcommand{\cmark}{\ding{51}}
\newcommand{\xmark}{\ding{55}}
\definecolor{loopsky}{RGB}{218,239,252}
\definecolor{secondsky}{RGB}{240,248,254}
\newlength{\resulttablewidth}
\makeatletter
\newcommand{\resultrow}[1]{\rlap{\hspace{-\tabcolsep}\color{#1}%
  \rule[-\dp\@arstrutbox]{\resulttablewidth}{\dimexpr\ht\@arstrutbox+\dp\@arstrutbox\relax}}}
\makeatother
\newcommand{\resulthead}[1]{\begin{tabular}[c]{@{}c@{}}#1\end{tabular}}

\title{LoopWAM: Closed-Loop Generalist Policy Improvement from Deployment Rollouts}

\author{%
  Wenxuan Guo\textsuperscript{1,2,3,*}\quad\quad
  Yingping Liang\textsuperscript{2,*}\quad\quad
  Zhaoguo Wang\textsuperscript{1,*}\quad\quad
  Ziyuan Li\textsuperscript{1}\quad\quad
  Angyuan Ma\textsuperscript{1}\newline
  Zhenyu Wu\textsuperscript{3}\quad\quad
  Chengkun Wang\textsuperscript{3}\quad\quad
  Bingyao Yu\textsuperscript{1}\quad\quad
  Ze Chen\textsuperscript{2}\quad\quad
  Borui Zhang\textsuperscript{1,3}\newline
  Xiuwei Xu\textsuperscript{3,$\dagger$}\quad\quad
  Meng Zhang\textsuperscript{2,$\dagger$}\quad\quad
  Erjin Zhou\textsuperscript{2}\quad\quad
  Jiwen Lu\textsuperscript{1,$\dagger$}\newline
  {\normalfont\fontsize{10}{14}\selectfont
   \textsuperscript{1}Tsinghua University\quad\quad
   \textsuperscript{2}Dexmal\quad\quad
   \textsuperscript{3}Loopeva}\newline
  {\normalfont\fontsize{8}{11}\selectfont
   * Core Contributors\qquad
   \textsuperscript{$\dagger$} Corresponding Authors}%
}
\hypersetup{pdfauthor={Wenxuan Guo, Yingping Liang, Zhaoguo Wang, Ziyuan Li, Angyuan Ma, Zhenyu Wu, Chengkun Wang, Bingyao Yu, Ze Chen, Borui Zhang, Xiuwei Xu, Meng Zhang, Erjin Zhou, Jiwen Lu}}

\begin{document}
\maketitle
\raggedbottom
\brokenpenalty=10000
\setlength{\emergencystretch}{3em}

\begin{abstract}

Scaling expert demonstrations alone does not enable generalist robot policies to improve from deployment.
Real-world rollouts contain successes, suboptimal behavior, and failures: imitating every recorded action can reinforce poor behavior, while filtering out failures discards evidence of physical interactions and their consequences.
We introduce \textbf{LoopWAM}, a closed-loop world-action model that learns directly from this full quality spectrum, using rollouts from both itself and other deployed policies.
Source-dependent noise sampling unifies behavior learning and action-conditioned world modeling, while quality steering and quality-weighted supervision retain full video supervision across qualities and regulate action imitation.
Video and action tokens interact throughout parallel denoising, allowing learned consequences to inform action generation directly.
On RoboChallenge Table30-V2, three deployment-and-training rounds raise success from 23.67\% to 43.33\%, a 19.7-percentage-point gain, with a final score of 56.80.
LoopWAM also achieves 94.5\% average success on RoboTwin~2.0 and 49.5\% overall success on RoboCasa365 without additional embodied pretraining, obtaining the best overall results among the compared methods across all three benchmarks.
On challenging shirt folding task, two rounds of deployment-driven improvement after single-task post-training raise success from 13.20\% to 55.56\%.
These results demonstrate closed-loop improvement from mixed-quality experience under both multi-task learning and single-task post-training. 
\href{https://loop-wam.github.io/}{\textcolor{orange}{Project Page}}
\end{abstract}
\abscontent

\section{Introduction}
\label{sec:introduction}

Scaling human-collected robot data has expanded the capabilities of generalist manipulation policies~\citep{o2024open}, but demonstration scaling alone does not close the loop between deployment and continual improvement.
Deployment exposes states, interactions, and failure modes beyond curated demonstrations, providing new experience for an embodied data flywheel.
Real-world reinforcement learning (RL)~\citep{xu2026rlt} can improve policy performance during deployment, but may narrow the model's predictive distribution and often struggles to learn multi-task generalist policies.
Although distilling a generalist policy from multiple RL specialists~\citep{xiao2025pld} can address this issue, the generalist's supervision comes from curated successful rollouts rather than the full interaction history.
Therefore, we aim to study direct generalist improvement from the full mixture of successful, suboptimal, and failed deployment rollouts.
The challenge is to use this experience productively: imitating every recorded action risks absorbing undesirable behavior, while retaining only high-quality trajectories discards informative interactions and their outcomes.

Recent VLA and world-action models address important pieces of this problem.
On the VLA side, $\pi_{0.6}$~\citep{physicalintelligence2025pi06} uses advantage-conditioned learning to incorporate autonomous experience and interventions, while $\pi_{0.7}$~\citep{physicalintelligence2026pi07} uses quality, speed, and mistake metadata to absorb mixed-quality data and distill specialist capabilities into a generalist.
These VLA methods use steering during training to model behavior distributions across data qualities and specify a desirable mode at inference for smooth execution, but rely on VLM-based language--vision--action representations without an explicit objective for learning the physical dynamics reflected in action--video transitions.
Consequently, suboptimal and failed rollouts provide behavior targets but do not directly supervise learning of their physical consequences.
World modeling offers a more elegant solution: WAMs~\citep{pai2025mimicvideo,ye2026world,lingbot-va2026} couple predictive visual representations with action generation, while action-conditioned world models (ACWMs)~\citep{guo2025ctrlworld,gao2026dreamdojo} predict future observations conditioned on actions.
Combining the two allows ACWM to learn action consequences from data of different qualities, while WAM learns appropriate actions from expert demonstrations.
$\tau_0$-WM~\citep{zhou2026tau0wm} trains separate WAM and ACWM, routing failure and recovery data to ACWM training and use ACWM as selector for policy improvement.
FACT~\citep{peng2026fact} implements the two roles within a shared model, but retains sequential action--consequence generation. It only supports success/failure supervision split, with failure consequences influencing the policy implicitly through shared representations.
These approaches expose a remaining design challenge in learning from mixed-quality deployment experience: VLA steering lacks natural action--video dependency modeling, while consequence-aware WAMs still exploit rollout data indirectly through an ACWM interface---a simulator for action selection or a subsequent consequence-prediction stage that influences the policy through shared representations.
Can a single WAM directly learn from mixed-quality deployment rollouts, with all qualities supervising the model and learned consequences participating directly in action generation?

\begin{figure}[t]
    \centering
    \includegraphics[width=\linewidth]{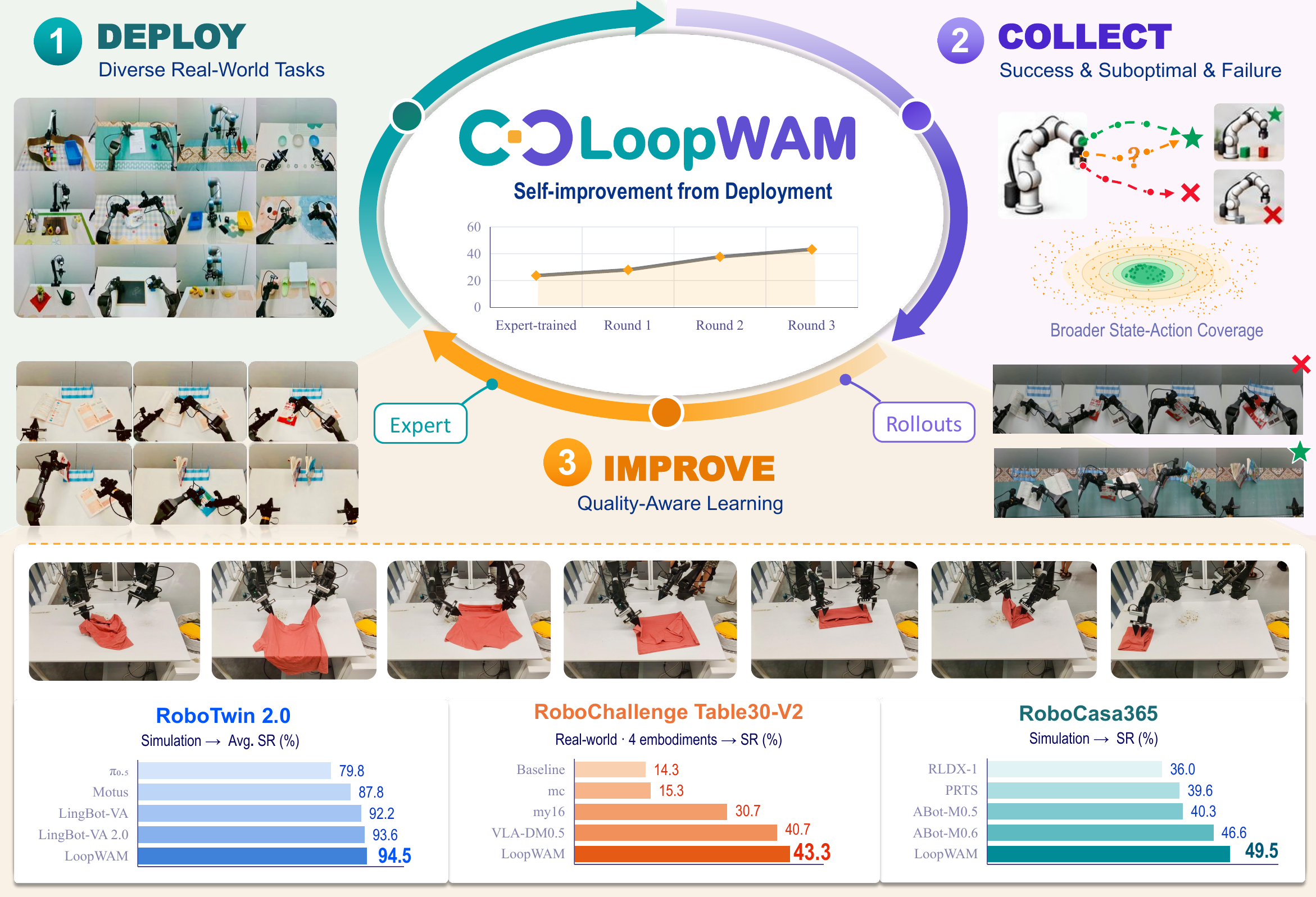}
    \caption{\textbf{LoopWAM learns from deployment.}
    Mixed-quality rollouts return to training through quality steering, source-dependent noise sampling, and quality-weighted supervision.
    Center: closed-loop self-improvement on RoboChallenge Table30-V2, where three rounds of deployment and retraining increase overall success from 23.67\% to 43.33\%.
    Bottom: benchmark comparisons.}
    \label{fig:teaser}
\end{figure}

We introduce \textbf{LoopWAM}, a closed-loop World-Action Model that updates a generalist directly from mixed-quality deployment rollouts.
Our core idea is to let one WAM directly absorb rollouts across the full quality spectrum and learn from each.
To achieve this, we first use \emph{source-dependent noise sampling} to unify WAM behavior learning and ACWM consequence learning within a single model and generation stage.
Video and action noise levels are sampled independently, allowing low-noise rollout actions to guide prediction of their observed consequences.
Within this unified model, \emph{quality steering} represents behaviors of different qualities and selects the highest-quality condition at deployment.
To control how strongly each behavior is imitated, we further apply \emph{quality-weighted supervision}, reducing imitation of suboptimal actions and masking failed-action targets while retaining video supervision across all qualities.
Together, these mechanisms let every trajectory supervise the joint model according to its behavioral quality and observed transitions.
Video and action tokens interact bidirectionally throughout one parallel denoising process, allowing learned future representations to participate in action formation without a separate consequence-evaluation stage.
We close the loop by repeatedly collecting rollouts and retraining the generalist on demonstrations and accumulated experience before redeployment.
On \textbf{RoboChallenge Table30-V2}~\citep{robochallenge2026table30v2}, three rounds raise overall success from \textbf{23.67\%} to \textbf{28.00\%}, \textbf{37.67\%}, and \textbf{43.33\%}, a \textbf{19.7-percentage-point improvement}, with a final score of \textbf{56.80}, ranking first in both success and score among the compared entries (Table~\ref{tab:robochallenge_results}).
LoopWAM achieves \textbf{94.5\%} average success on \textbf{RoboTwin~2.0}~\citep{chen2025robotwin2}, exceeding the reported 87.5\% of FACT and 93.6\% of LingBot-VA~2.0, and \textbf{49.5\%} on the 50-task \textbf{RoboCasa365} evaluation~\citep{nasiriany2026robocasa365}, above ABot-M0.6's 46.6\% (Tables~\ref{tab:robotwin_results}--\ref{tab:robocasa_results}).
We further evaluate deployment-driven self-improvement after single-task post-training on challenging shirt folding, which involves deformable objects and multiple coordinated steps.
Two deployment-and-training rounds raise success from \textbf{13.20\%} to \textbf{55.56\%} (Section~\ref{sec:closed_loop_results}).
These results establish that directly learning from mixed-quality rollouts with parallel world--action denoising can deliver leading benchmark performance and repeated deployment improvement in both multi-task and single-task settings.

\section{Related Work}
\label{sec:related_work}

\textbf{Robot Manipulation Policies.}
Robot manipulation policies increasingly combine scalable imitation learning with pretrained representations.
ACT~\citep{zhao2023learning} and Diffusion Policy~\citep{chi2025diffusion} learn action sequences from demonstrations, while RT-2~\citep{zitkovich2023rt2} and OpenVLA~\citep{kim2024openvla} transfer vision-language priors to language-conditioned control.
Flow-based policies such as $\pi_0$~\citep{black2024pi0} and $\pi_{0.5}$~\citep{physicalintelligence2025pi05} further scale continuous action generation and generalization.
Open X-Embodiment~\citep{o2024open} and DROID~\citep{khazatsky2024droid} expand training coverage across embodiments and environments.
Beyond offline demonstrations, DAgger~\citep{ross2011dagger} addresses policy-induced distribution shift by collecting expert labels on visited states.
LoopWAM studies learning from the recorded observations and actions of deployment rollouts, using trajectory quality to guide policy updates while retaining world supervision from unsuccessful interactions.

\textbf{World Action Models.}
Video-based robot policies use predictive visual representations to support control.
mimic-video~\citep{pai2025mimicvideo} conditions an action decoder on representations from a pretrained video model.
DreamZero~\citep{ye2026world} jointly generates video and actions, while LingBot-VA~\citep{lingbot-va2026} organizes visual prediction and action decoding within a causal autoregressive framework.
LingBot-VA~2.0~\citep{zhang2026lingbotva2} further develops native video-action pretraining with a control-oriented tokenizer.
Efficiency-oriented designs include Fast-WAM~\citep{yuan2026fast}, which retains video co-training but removes future-video generation at inference, and GigaWorld-Policy~\citep{ye2026gigaworld}, which learns action-conditioned visual dynamics while allowing action-only inference.
Being-H0.7~\citep{luo2026beingh07} instead uses future observations to supervise latent representations.
These studies establish several ways for predictive supervision to support action learning.
LoopWAM focuses on closing the deployment--training loop: mixed-quality rollouts broaden world-model training, while joint video--action modeling, quality steering, and weighted action supervision connect consequence awareness with high-quality behavior.

\section{Method}
\label{sec:method}

\begin{figure}[t]
    \centering
    \includegraphics[width=\linewidth,pagebox=cropbox]{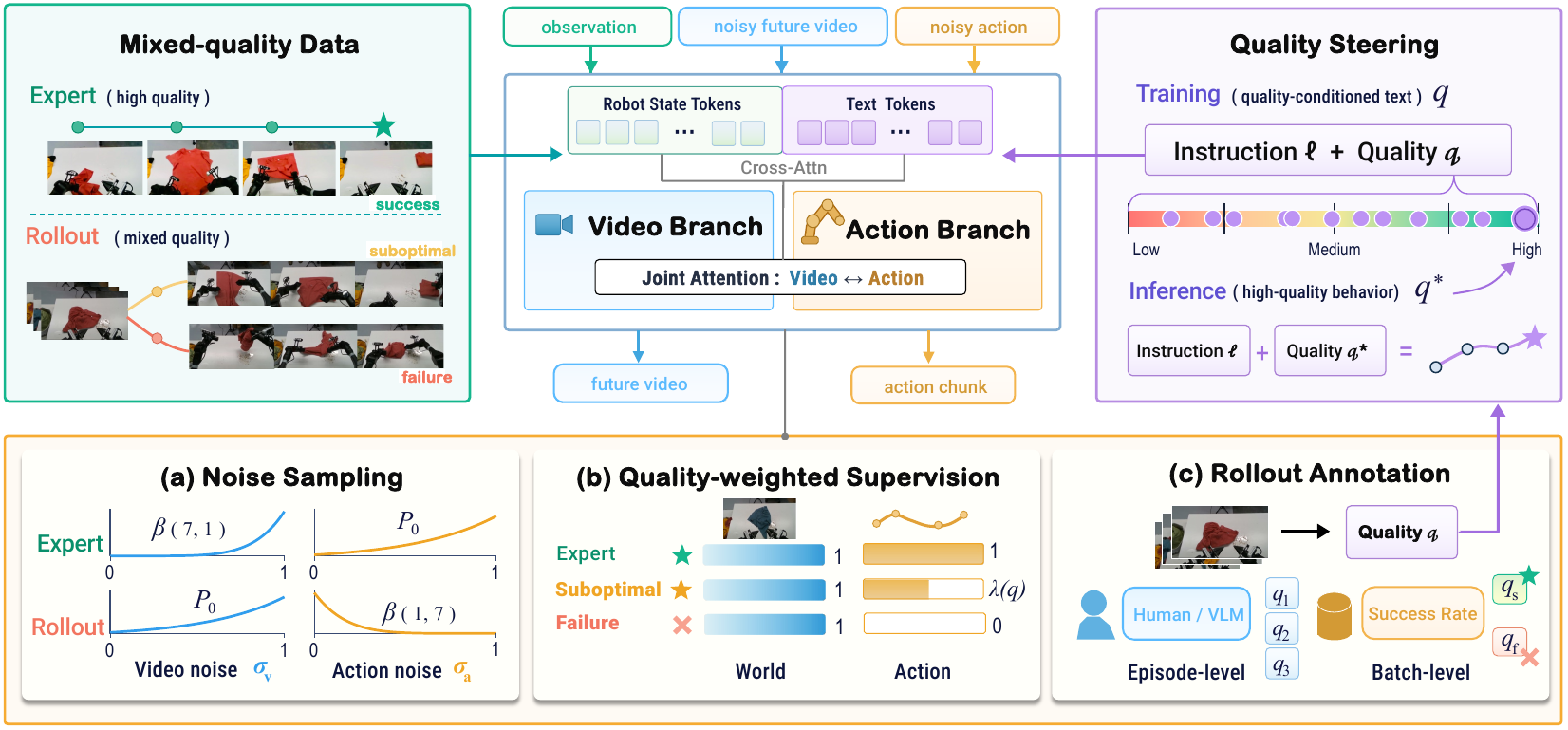}
    \caption{\textbf{Overview of LoopWAM.}
    LoopWAM is structured as a Mixture-of-Transformer (MoT) architecture, where bidirectional attention enables feature interaction between the video and action branches.
    Training combines quality-conditioned instructions, source-dependent noise sampling, and quality-weighted action supervision.
    Expert trajectories supervise both modalities, while lower-quality rollouts retain full world supervision and contribute to action imitation according to their quality.
    The trained model is deployed using the expert score $q^\star$, and recorded experience returns to the next training round.}
    \label{fig:method}
\end{figure}

LoopWAM learns a world--action model that can be updated with experience collected during deployment.
The framework couples video and action generation within a shared denoising process, then uses trajectory quality to distinguish how deployment data supervise world modeling and control.
Figure~\ref{fig:method} summarizes the architecture and training procedure.
We first introduce joint world--action modeling, followed by quality-aware learning from mixed trajectories and the deployment--learning loop.

\subsection{Joint World--Action Modeling}
\label{sec:coupled_model}

We consider language-conditioned manipulation with visual observation $o_t$, robot state $s_t$, and instruction $\ell$.
The prediction targets are an action chunk $A_t=(a_t,\ldots,a_{t+H-1})$ and the corresponding future observations $V_t$, sampled over the same horizon.
A pretrained video VAE maps $(o_t,V_t)$ to an observation latent $z_o$ and future latents $Z_t$.
LoopWAM models their joint distribution
\begin{equation}
    p_\theta(A_t,Z_t\mid o_t,s_t,\ell,q),
    \label{eq:quality_joint}
\end{equation}
where $q$ is a numerical score of trajectory quality, as detailed in Section~\ref{sec:quality_steering}.
This unified formulation enables a single model to learn action generation and visual dynamics from paired action--observation sequences~\citep{zhu2025unifiedworldmodelscoupling}.

We instantiate this formulation with a dual-stream diffusion transformer whose video backbone is initialized from Wan2.2-TI2V-5B~\citep{wan2025}.
The video and action streams have separate token projections, transformer parameters, and prediction heads, with joint attention connecting the streams at every layer.
To train them with flow matching~\citep{lipman2023flow}, we independently perturb each target $X_m$, where $X_v=Z_t$ and $X_a=A_t$:
\begin{equation}
    X_m^{\sigma_m}=(1-\sigma_m)X_m+\sigma_m\epsilon_m,
    \qquad \epsilon_m\sim\mathcal N(0,I),\quad m\in\{v,a\}.
    \label{eq:flow_path}
\end{equation}
The joint network predicts velocity fields $(v_\theta^v,v_\theta^a)$ with targets $u_m=\epsilon_m-X_m$.
Each stream receives noise level $\sigma_m$, with $0$ denoting clean data and $1$ pure noise; the observation latent $z_o$ remains clean.

The attention structure couples the two predictions throughout denoising.
Denote observation, future-video, and action tokens by $O$, $V$, and $A$, respectively.
Their block mask, with queries indexing rows and keys indexing columns, is
\begin{equation}
    M_{(O,V,A)}=
    \begin{pmatrix}
        1 & 0 & 0\\
        1 & 1 & 1\\
        1 & 1 & 1
    \end{pmatrix},
    \label{eq:joint_mask}
\end{equation}
where $1$ permits attention.
Observation tokens attend only within the initial frame.
Both streams cross-attend to the quality-conditioned instruction; robot-state tokens additionally condition the action stream.
Together with the training scheme in Section~\ref{sec:quality_steering}, this architecture supports learning both embodied world representations and awareness of action consequences.
Dense video supervision on expert and mixed-quality rollouts broadens the interactions from which these representations are learned.
Through bidirectional attention, the world-prediction loss also trains features used for action generation.
This training couples learning how tasks can be completed with learning how particular actions affect their outcomes, encouraging consistency between video and action generation.

\subsection{Quality-Aware Learning from Mixed Trajectories}
\label{sec:quality_steering}

Our training corpus combines expert demonstrations $\mathcal D_E$ with rollout data $\mathcal D_R$ of arbitrary quality, including successful, suboptimal, and failed executions.
These rollouts substantially broaden the state--action--outcome distribution beyond expert paths, providing world-model supervision even when their actions are undesirable imitation targets.
Each trajectory is assigned a source indicator $d\in\{E,R\}$; every training segment extracted from that trajectory uses the same indicator.
We combine source-dependent noise sampling, quality conditioning, and selective action supervision to learn from this broader experience while favoring high-quality behavior.

\textbf{Source-dependent noise sampling.}
\label{sec:noise_sampling}
Jointly predicting video and actions in a WAM does not by itself ensure that the predicted futures reflect the consequences of the generated actions.
To improve consistency between video and action generation and make world predictions more useful for control, we aim to strengthen action-conditioned world modeling (ACWM) within the same unified model.
We therefore use source-dependent noise sampling to make recorded rollout actions accessible when predicting their observed consequences, while retaining joint video--action generation in a single training stage.
Specifically, we select different noise distributions for expert and rollout segments, while drawing video and action noise levels independently within each segment:
\begin{equation}
\begin{aligned}
    d=E:&\quad \sigma_v\sim\operatorname{Beta}(7,1),
                 &&\sigma_a\sim\mathcal P_0,\\
    d=R:&\quad \sigma_v\sim\mathcal P_0,
                 &&\sigma_a\sim\operatorname{Beta}(1,7).
\end{aligned}
\label{eq:source_noise}
\end{equation}
Here $\mathcal P_0$ is the base noise distribution obtained by sampling $u\sim\mathcal U(0,1)$ and setting $\sigma=\phi_\kappa(u)=\kappa u/[1+(\kappa-1)u]$, with $\kappa=5$.
The Beta distributions~\citep{johnson1995continuous} are applied directly to $\sigma$.
For expert data, high video noise emphasizes recovering expert futures from strongly corrupted inputs, alongside action generation over the base noise distribution.
For rollouts, low action noise preserves much of the recorded motion while the video stream predicts its visual consequences.
Training across these executions broadens the action--outcome distribution seen by the world model, including transitions in which the task remains incomplete or an interaction fails.
Joint attention then connects this consequence-learning signal to the representations used for action generation.

\textbf{Quality steering.}
Within this unified training scheme, each trajectory is assigned a numerical quality score $q$, which is shared by all training segments extracted from it.
We encode quality as a suffix to the instruction, $\ell_q=[\ell;\texttt{Quality:}\ q]$, and use its text embeddings to condition both video and action streams.
We support two trajectory annotation schemes: (1) \emph{Episode-level annotation}, where human annotators or a vision-language model (VLM) score each trajectory individually for fine-grained quality estimation; and (2) \emph{coarse-level annotation}, where a single human-provided score is assigned to all rollouts from a policy checkpoint based on its overall success rate and motion smoothness.
Under coarse annotation, successful trajectories receive the checkpoint score while failed trajectories receive the minimum score, requiring only one annotation per checkpoint.
Both schemes represent rollout quality using numerical scores.
For expert demonstrations, we assign an expert score $q^\star$ that is higher than the maximum rollout quality score.
The quality-conditioned instruction allows different behaviors under the same task instruction to be represented within one model.
During deployment, we condition on $q=q^\star$ to steer sampling toward expert-quality behavior.

\textbf{Quality-weighted supervision.}
\label{sec:quality_objective}
Given these noise levels and quality scores, we retain video prediction on all trajectory qualities and modulate action supervision with a nondecreasing weight $\lambda(q)\in[0,1]$.
Expert trajectories receive full action supervision with $\lambda(q^\star)=1$, while failed trajectories use $\lambda(q)=0$ and receive no action supervision.
For sample $i$ and modality $m$, define the denoising loss
\begin{equation}
    \ell_{m,i}=\omega(\sigma_{m,i})
    \left\|v_{\theta,i}^{m}-(\epsilon_{m,i}-X_{m,i})\right\|_{\mathrm{valid}}^2,
    \label{eq:modality_loss}
\end{equation}
where the norm denotes mean squared error over valid target elements and $\omega$ is the common timestep weighting function.
The clean observation latent is excluded from video supervision, and padded targets are masked in both modalities.
For a minibatch of $B$ segments, the complete objective is
\begin{equation}
    \mathcal L_{\mathrm{LoopWAM}}
    =\underbrace{\frac{1}{B}\sum_{i=1}^{B}\ell_{v,i}}_{\text{world supervision}}
    +\underbrace{\frac{\sum_{i=1}^{B}\lambda(q_i)\ell_{a,i}}
        {\max\!\left(1,\sum_{i=1}^{B}\lambda(q_i)\right)}}_{\text{quality-weighted action supervision}}.
    \label{eq:quality_loss}
\end{equation}
The source indicator determines corruption, whereas quality controls the behavior condition and action weight.
In particular, $\lambda(q)=0$ removes direct imitation of a trajectory's actions while preserving world-model gradients through the jointly attended video and action representations.
Quality steering and weighted supervision thus allow broader transition learning while concentrating action imitation on more desirable behavior.
The data preprocessing are specified in Appendix~\ref{app:implementation}.

\subsection{Closed-Loop Self-Improvement}
\label{sec:deployment_loop}

The deployed policy uses the high-quality action marginal of the joint model:
\begin{equation}
    \pi_\theta(A_t\mid o_t,s_t,\ell;q^\star)
    =\int p_\theta(A_t,Z_t\mid o_t,s_t,\ell,q^\star)\,\mathrm d Z_t.
    \label{eq:steered_policy}
\end{equation}
In practice, we sample this marginal by jointly denoising video and actions from Gaussian noise, holding the current observation latent fixed.
Starting from an expert-trained model, we collect and score deployment rollouts, then update the model on expert demonstrations and all accumulated rollouts using Equation~\ref{eq:quality_loss}.
Each updated model is redeployed with the expert score $q^\star$ to continue the deployment--learning loop.

More broadly, LoopWAM's self-improvement updates can use rollouts from arbitrary deployed policies, even poorly performing ones.
The joint architecture and quality-aware objective accommodate their varying trajectory qualities, so each trajectory contributes according to its observed transitions and behavioral quality.
This extends self-improvement to learning from deployment experience collected by other policies as well as LoopWAM itself.
We validate this broader setting on RoboTwin~2.0 and RoboCasa365 (Section~\ref{sec:benchmark_results}).

\section{Experiments}
\label{sec:experiments}

\subsection{Implementation Details}
\label{sec:implementation}

We initialize the video backbone, text encoder, and video VAE from Wan2.2-TI2V-5B~\citep{wan2025}, and use an action DiT with a hidden dimension of 1024.
Each training segment pairs a 64-step action chunk with nine RGB frames sampled at intervals of eight control steps, including the initial observation.
Multi-camera views are stitched into a single image, and actions and robot states are normalized using expert-data statistics.
Training follows the quality-conditioned joint objective in Section~\ref{sec:quality_steering}, and deployment uses the expert score $q^\star$.
Table30-V2 uses relative joint targets and absolute gripper commands.
Additional training, preprocessing, and subtask-segmentation details are provided in Appendix~\ref{app:implementation}.

\subsection{Benchmark Experimental Setup}
\label{sec:benchmark_setup}

We evaluate LoopWAM on the real-robot benchmark RoboChallenge Table30-V2 and two simulation benchmarks, RoboTwin~2.0 and RoboCasa365.
RoboChallenge evaluates closed-loop self-improvement using LoopWAM's own real-robot deployments.
The two simulation benchmarks additionally test learning from different deployed policies by augmenting expert demonstrations with rollouts from both expert-trained LoopWAM and others.
Across these settings, successful, suboptimal, and failed rollouts expand the state--action--outcome distribution beyond expert demonstrations.

\paragraph{RoboChallenge Table30-V2.}
RoboChallenge Table30-V2~\citep{robochallenge2026table30v2} is a standardized real-robot benchmark comprising 30 tabletop manipulation tasks across four embodiments: ALOHA, DOS-W1, ARX5, and UR5.
Its tasks test capabilities including tool use, precise manipulation, and coordinated bimanual control.
We follow the official multi-task evaluation protocol and report overall performance across embodiments.
Starting from one expert-trained multi-task LoopWAM per embodiment, we conduct three successive rounds of closed-loop self-improvement.
In each deployment round, we run each model on its assigned tasks and collect about 100 rollout trajectories per task.
These trajectories are quality-annotated and combined with expert demonstrations and rollout trajectories from all previous deployment rounds for further training.

\paragraph{RoboTwin~2.0.}
RoboTwin~2.0~\citep{chen2025robotwin2} comprises 50 bimanual manipulation tasks evaluated in clean and domain-randomized scenes.
Following the multi-task setup of LingBot-VA~\citep{lingbot-va2026}, we use 2,500 expert demonstrations from clean scenes and 25,000 from heavily randomized scenes.
We augment these demonstrations with rollouts generated by expert-trained LoopWAM, X-VLA~\citep{zheng2025xvla}, and $\pi_{0.5}$~\citep{physicalintelligence2025pi05}.
For each task, each policy contributes 20 clean and 130 randomized rollouts, yielding 22,500 trajectories across all 50 tasks and the three policies.
The combined rollout corpus has approximately the same total duration as the expert corpus.
Rollout collection and evaluation use different random seeds to avoid data leakage.
Rollout quality is annotated using the coarse-level numerical scores in Section~\ref{sec:quality_steering}.

\begin{table}[t]
    \centering
    \caption{Performance on RoboChallenge Table30-V2. We report per-embodiment success rates (SR, \%) and overall performance across four robot embodiments. Baselines are sorted by overall SR. Bold denotes the best performance, while blue and light blue highlight LoopWAM and the runner-up, respectively.}
    \label{tab:robochallenge_results}
    \footnotesize
    \setlength{\tabcolsep}{2pt}
    \renewcommand{\arraystretch}{1.12}
    \setlength{\resulttablewidth}{\linewidth}
    \begin{tabular*}{\resulttablewidth}{l@{\extracolsep{\fill}}*{10}{r}}
        \toprule
        \raisebox{-0.75\normalbaselineskip}[0pt][0pt]{Method} & \multicolumn{2}{c}{ALOHA} & \multicolumn{2}{c}{DOS-W1} & \multicolumn{2}{c}{ARX5} & \multicolumn{2}{c}{UR5} & \multicolumn{2}{c}{Overall} \\
        \cmidrule(lr){2-3}\cmidrule(lr){4-5}\cmidrule(lr){6-7}\cmidrule(lr){8-9}\cmidrule(lr){10-11}
         & Score & SR & Score & SR & Score & SR & Score & SR & Score & SR \\
        \midrule
        debuggers\_vla & 5.40 & 0.00 & 5.15 & 2.00 & 53.21 & 34.29 & 27.83 & 23.33 & 18.72 & 11.00 \\
        W0 & 12.25 & 2.00 & 4.40 & 0.00 & 48.64 & 30.00 & 49.50 & 43.33 & 21.85 & 12.00 \\
        test\_01 & 6.25 & 1.00 & 8.95 & 3.00 & 49.50 & 28.57 & 61.17 & 43.33 & 22.73 & 12.33 \\
        holobrain & 11.75 & 3.00 & 13.30 & 6.00 & 53.64 & 30.00 & 70.67 & 43.33 & 27.93 & 14.33 \\
        Baseline Model & 21.55 & 6.00 & 17.35 & 6.00 & 60.14 & 32.86 & 44.83 & 26.67 & 31.48 & 14.33 \\
        MagicBot & 4.70 & 0.00 & 6.85 & 5.00 & 47.57 & 32.86 & 58.67 & 53.33 & 20.82 & 14.67 \\
        mc & 10.60 & 7.00 & 2.00 & 0.00 & 73.64 & 47.14 & 22.50 & 20.00 & 23.63 & 15.33 \\
        JV0 & 12.70 & 5.00 & 17.80 & 16.00 & 49.64 & 20.00 & 58.33 & 43.33 & 27.58 & 16.00 \\
        my16 & 18.95 & 13.00 & 36.60 & 27.00 & 73.29 & 54.29 & 56.50 & 46.67 & 41.27 & 30.67 \\
        \resultrow{secondsky}VLA-DM0.5 & 26.50 & 18.00 & 55.45 & 42.00 & 79.07 & 58.57 & 86.50 & 70.00 & 54.42 & 40.67 \\
        \midrule
        LoopWAM (expert-trained) & 23.05 & 8.00 & 46.35 & 38.00 & 37.00 & 24.29 & 43.33 & 26.67 & 36.10 & 23.67 \\
        \resultrow{loopsky}\textbf{LoopWAM} & 38.65 & 26.00 & 54.90 & 42.00 & 76.21 & 55.71 & 78.33 & 76.67 & \textbf{56.80} & \textbf{43.33} \\
        \bottomrule
    \end{tabular*}
\end{table}

\begin{table}[t]
    \centering
    \caption{Performance on RoboTwin~2.0. We report success rates (SR, \%) under clean and randomized evaluation settings. Embodied PT. denotes additional embodied-data pretraining before benchmark training. LoopWAM is trained with the mixed expert--rollout corpus described in Section~\ref{sec:benchmark_setup}. Blue and light blue highlight LoopWAM and the runner-up, respectively.}
    \label{tab:robotwin_results}
    \footnotesize
    \setlength{\tabcolsep}{4pt}
    \renewcommand{\arraystretch}{1.05}
    \setlength{\resulttablewidth}{0.85\linewidth}
    \begin{tabular*}{\resulttablewidth}{l@{\extracolsep{\fill}}cccc}
        \toprule
        Method & Embodied PT. & Clean & Rand. & Average \\
        \midrule
        $\pi_0$~\citep{black2024pi0} & \cmark & 65.9 & 58.4 & 62.2 \\
        X-VLA~\citep{zheng2025xvla} & \cmark & 72.9 & 72.8 & 72.9 \\
        $\pi_{0.5}$~\citep{physicalintelligence2025pi05} & \cmark & 82.7 & 76.8 & 79.8 \\
        ABot-M0~\citep{yang2026abot} & \cmark & 81.2 & 80.4 & 80.8 \\
        Motus (from Wan2.2)~\citep{bi2025motus} & \xmark & 77.6 & 77.0 & 77.3 \\
        Motus~\citep{bi2025motus} & \cmark & 88.7 & 87.0 & 87.8 \\
        LingBot-VA~\citep{lingbot-va2026} & \cmark & 92.9 & 91.5 & 92.2 \\
        Fast-WAM~\citep{yuan2026fast} & \xmark & 91.9 & 91.8 & 91.8 \\
        AHA-WAM~\citep{cai2026ahawam} & \xmark & 93.4 & 92.2 & 92.8 \\
        \resultrow{secondsky}LingBot-VA~2.0~\citep{zhang2026lingbotva2} & \cmark & 93.8 & 93.4 & 93.6 \\
        \midrule
        \resultrow{loopsky}\textbf{LoopWAM} & \xmark & \textbf{94.6} & \textbf{94.4} & \textbf{94.5} \\
        \bottomrule
    \end{tabular*}
\end{table}

\begin{table}[t]
    \centering
    \caption{Performance on RoboCasa365. We report success rates (SR, \%) across three evaluation splits. Baseline results are obtained from the official leaderboard. Embodied PT. follows the definition in Table~\ref{tab:robotwin_results}. Blue and light blue highlight LoopWAM and the runner-up, respectively.}
    \label{tab:robocasa_results}
    \footnotesize
    \setlength{\tabcolsep}{3pt}
    \renewcommand{\arraystretch}{1.05}
    \setlength{\resulttablewidth}{0.97\linewidth}
    \begin{tabular*}{\resulttablewidth}{l@{\extracolsep{\fill}}ccccc}
        \toprule
        Method & \resulthead{Embodied PT.} & \resulthead{Atomic Seen} & \resulthead{Composite Seen} & \resulthead{Composite Unseen} & Overall \\
        \midrule
        Diffusion Policy~\citep{chi2025diffusion} & \xmark & 15.7 & 0.2 & 1.3 & 6.1 \\
        $\pi_0$~\citep{black2024pi0} & \cmark & 34.6 & 6.1 & 1.1 & 14.8 \\
        $\pi_{0.5}$~\citep{physicalintelligence2025pi05} & \cmark & 39.6 & 7.1 & 1.2 & 16.9 \\
        GigaWorld-Policy~0.1~\citep{ye2026gigaworld} & \cmark & 44.4 & 11.8 & 2.9 & 20.7 \\
        GR00T N1.6~\citep{nvidia2025gr00t} & \cmark & 51.1 & 9.4 & 1.7 & 21.9 \\
        GR00T N1.5~\citep{nvidia2025gr00t} & \cmark & 50.7 & 14.8 & 2.7 & 23.9 \\
        RLDX-1~\citep{kim2026rldx1} & \cmark & 67.6 & 27.9 & 8.5 & 36.0 \\
        PRTS~\citep{zhang2026prts} & \cmark & 66.3 & 30.3 & \textbf{18.8} & 39.6 \\
        ABot-M0.5~\citep{chen2026abot05} & \cmark & 75.6 & 37.7 & 3.3 & 40.3 \\
        \resultrow{secondsky}ABot-M0.6~\citep{yang2026abot} & \cmark & 79.4 & 48.3 & 7.9 & 46.6 \\
        \midrule
        \resultrow{loopsky}\textbf{LoopWAM} & \xmark & \textbf{79.7} & \textbf{50.3} & 14.9 & \textbf{49.5} \\
        \bottomrule
    \end{tabular*}
\end{table}

\paragraph{RoboCasa365.}
RoboCasa365~\citep{nasiriany2026robocasa365} is a large-scale simulation framework for household manipulation in diverse kitchen environments.
We evaluate on its 50-task benchmark, which includes atomic and composite manipulation tasks.
Training combines expert demonstrations with rollouts generated by expert-trained LoopWAM and RLDX-1~\citep{kim2026rldx1}, with the combined rollout duration approximately matching that of the expert data.
Both expert and rollout data follow the Human300 training split: the 16 Composite-Unseen tasks are excluded from training and evaluated zero-shot.
As in RoboTwin~2.0, training and evaluation data use different random seeds.

\subsection{Benchmark Results}
\label{sec:benchmark_results}

\paragraph{RoboChallenge Table30-V2.}
Table~\ref{tab:robochallenge_results} compares expert-trained and final LoopWAM with leading entries from the official leaderboard~\citep{robochallenge2026leaderboard}.
After three deployment-and-training rounds, LoopWAM attains an overall score of 56.80 and 43.33\% success, exceeding VLA-DM0.5 by 2.38 score points and 2.67\% in success, and ranking first on both metrics among the compared methods.
Appendix~\ref{app:leaderboard} includes a leaderboard snapshot.
Section~\ref{sec:closed_loop_results} analyzes the successive deployment updates that produce this improvement over expert-only training.

\paragraph{RoboTwin~2.0.}
LoopWAM achieves 94.58\% success in clean scenes and 94.42\% under domain randomization, averaging 94.50\% (Table~\ref{tab:robotwin_results}).
Without additional embodied pretraining, it improves over AHA-WAM by 1.70 percentage points and surpasses the pretrained LingBot-VA~2.0 by 0.90 points on average.
Together, these results show that joint learning from expert demonstrations and diverse deployment rollouts yields a competitive multi-task policy under both evaluation settings.

\paragraph{RoboCasa365.}
LoopWAM achieves 79.7\%, 50.3\%, and 14.9\% success on Atomic-Seen, Composite-Seen, and Composite-Unseen, respectively, yielding 49.5\% overall success across the 50 tasks (Table~\ref{tab:robocasa_results}).
It exceeds ABot-M0.6 by 2.9 percentage points overall without additional embodied pretraining.
The improvement extends to held-out composite tasks, demonstrating transfer to task compositions absent from both expert and rollout training data.

\subsection{Closed-Loop Improvement from Deployment}
\label{sec:closed_loop_results}

\begin{figure}[t]
    \centering
    \includegraphics[width=\linewidth]{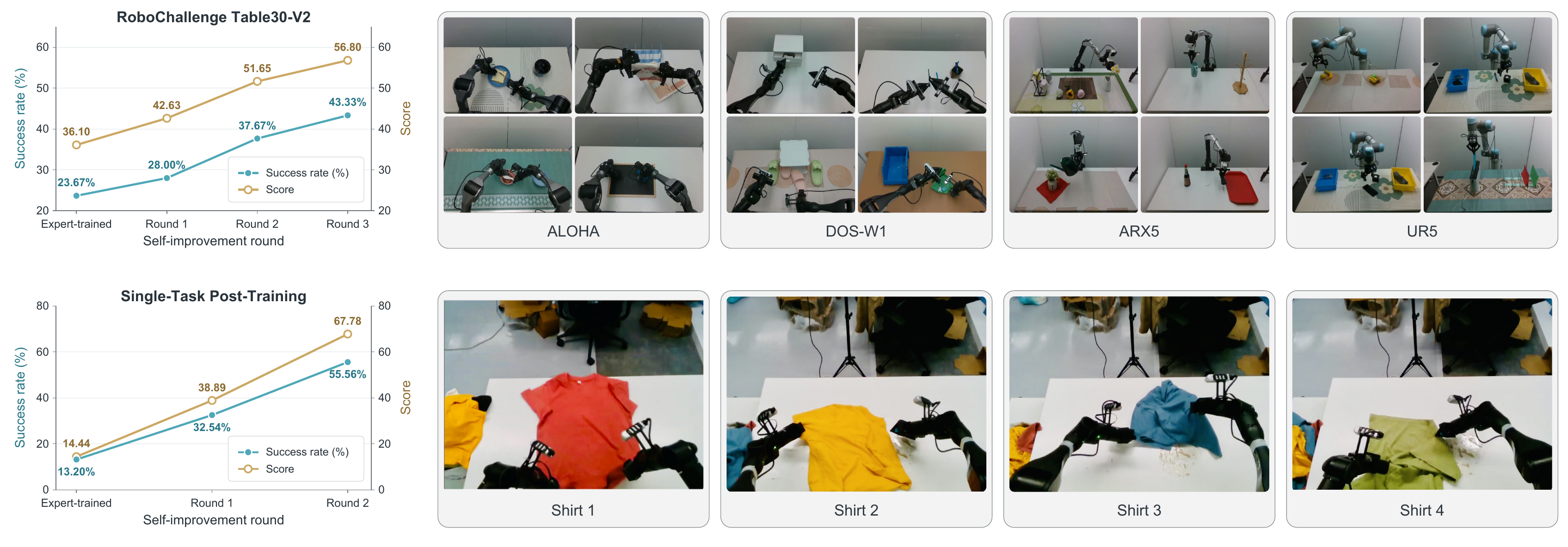}
    \caption{Closed-loop improvement from real-robot deployment. 
    Top: three deployment-and-training iterations on RoboChallenge Table30-V2, covering 30 tasks across four robot embodiments. 
    Bottom: two deployment-and-training iterations on shirt folding after task-specific expert post-training of the DOS-W1 model. 
    Each plot reports success rate and score over successive improvement rounds.}
    \label{fig:closed_loop_improvement}
\end{figure}

We examine deployment-driven improvement in two real-robot settings: multi-task learning and single-task post-training (Figure~\ref{fig:closed_loop_improvement}).
Both follow the update procedure in Section~\ref{sec:deployment_loop}: At each iteration, we deploy the current policy and assign quality scores to the collected rollouts, then update the model through quality-aware joint training on expert demonstrations and all rollouts collected so far.

\textbf{Multi-task deployment.}
As in Section~\ref{sec:benchmark_setup}, we train multi-task models for ALOHA, DOS-W1, ARX5, and UR5, covering the 30 Table30-V2 tasks.
Starting from expert-trained checkpoints, each model undergoes three rounds of deployment and training using experience from its assigned tasks.
Overall success increases from 23.67\% to 28.00\%, 37.67\%, and 43.33\%, while the scores rise from 36.10 to 42.63, 51.65, and 56.80.
Both aggregate metrics improve at every iteration, yielding a 19.7-percentage-point success gain over the initial deployment.
This demonstrates that LoopWAM can use its own deployment experience to improve policies trained across multiple tasks.

\textbf{Single-task post-training.}
We further evaluate shirt folding, a challenging deformable-object manipulation task with diverse and disordered initial configurations.
A shirt is casually tossed into the robot's workspace, creating varied and messy initial states in which its corners may be hidden.
The robot must identify suitable grasp points and progressively unfold the shirt until its corners are exposed, then grasp these corners to complete the subsequent folds.
We initialize from the DOS-W1 multi-task model trained on Table30-V2 and post-train it on newly collected expert demonstrations of shirt folding.
The resulting task-specific expert-trained checkpoint serves as the starting point for two rounds of deployment and self-improvement.
Success rises from 13.20\% to 32.54\% and 55.56\%, while the scores increase from 14.44 to 38.89 and 67.78.
These gains show that deployment feedback remains effective after adapting a multi-task model to a challenging individual task, supporting closed-loop improvement under both training regimes.

\subsection{Ablation Study}
\label{sec:ablation}

We use the RoboTwin~2.0 setup in Section~\ref{sec:benchmark_setup}, varying only quality steering and rollout action supervision; expert action supervision and video supervision remain unchanged.

\begin{table*}[t]
\caption{Ablation study on RoboTwin~2.0. We evaluate quality steering with different annotation schemes and assess the effect of quality-weighted rollout action supervision (Act.).}
\label{tab:robotwin_ablation}
\centering
\small
\setlength{\tabcolsep}{4pt}
\renewcommand{\arraystretch}{1.0}
\setlength{\resulttablewidth}{0.75\linewidth}
\begin{tabular*}{\resulttablewidth}{l@{\extracolsep{\fill}}cccc}
    \toprule
    Quality Steering & Quality-weighted Act. & Clean & Rand. & Average \\
    \midrule
    None & \xmark & 73.5 & 71.1 & 72.3 \\
    Extra-coarse & \xmark & 92.2 & 93.3 & 92.8 \\
    \resultrow{secondsky}Extra-coarse & \cmark & 93.9 & 93.3 & 93.6 \\
    Coarse & \xmark & 93.2 & 92.2 & 92.7 \\
    \midrule
    \resultrow{loopsky}\textbf{Coarse (ours)} & \cmark & \textbf{94.6} & \textbf{94.4} & \textbf{94.5} \\
    \bottomrule
\end{tabular*}
\end{table*}

Table~\ref{tab:robotwin_ablation} compares no steering, extra-coarse annotation, and our coarse-level annotation (Section~\ref{sec:quality_steering}).
Extra-coarse annotation distinguishes rollouts from all policies solely by success or failure: all failed rollouts are assigned \textit{Quality: low}, all successful rollouts are assigned \textit{Quality: medium}, and expert demonstrations are assigned \textit{Quality: high}.
In our coarse-level scheme, successful rollouts from the same policy are assigned a shared numerical score, which can differ across policies according to their overall success rates across tasks and manually assessed motion smoothness.
Failed rollouts are assigned the minimum score, while expert demonstrations are assigned $q^\star$, above the rollout score range. The ablation results reveal three key findings: 
\textbf{(1)} Quality steering is key to learning from mixed-quality deployment rollouts, enabling the model to distinguish trajectories of different quality.
\textbf{(2)} Even extra-coarse labels remain effective, demonstrating that fine-grained annotation is not required to benefit from steering.
\textbf{(3)} Quality-aware rollout action supervision further improves overall performance under both annotation schemes.

\section{Conclusion}
\label{sec:conclusion}

We presented LoopWAM for improving generalist world-action policies directly from deployment experience.
The key is to learn physical consequences from trajectories across the quality spectrum while controlling how their actions supervise behavior.
Source-dependent noise sampling, quality steering, and quality-weighted supervision implement this distinction within joint video--action denoising, allowing consequence representations to participate directly in action generation.
The same formulation accepts rollouts from both LoopWAM and other deployed policies. Experiments show the best overall results: a score of 56.80 and 43.33\% success on RoboChallenge Table30-V2, 94.5\% average success on RoboTwin~2.0, and 49.5\% overall success on RoboCasa365.
Repeated real-robot deployment raises Table30-V2 success by 19.7 percentage points over three rounds and improves shirt-folding success from 13.20\% to 55.56\% over two rounds after single-task post-training.
These findings support using mixed-quality interaction history as a training resource for continued improvement of both multi-task policies and policies adapted to individual tasks.

\bibliographystyle{plainnat}
\bibliography{references}

\clearpage
\appendix

\clearpage
\appendix

\section{Training and Data Processing Details}
\label{app:implementation}

\subsection{Trajectory-Quality Annotations}
\label{app:quality_annotations}

We use two numerical quality scales according to annotation granularity.
\emph{Episode-level annotation} uses a 1--12 scale, with rollout scores in $[1,10]$ and expert demonstrations assigned $q^\star=12$.
\emph{Coarse-level annotation} uses a 1--5 scale, with rollout scores in $\{1,2,3,4\}$ and expert demonstrations assigned $q^\star=5$.
In all settings, every failed trajectory receives $q=1$.
Expert demonstrations are treated as optimal reference trajectories and assigned scores above the rollout range; the same expert score is used at deployment.
Each score is assigned to a complete trajectory and inherited by all training segments extracted from it.

\textbf{RoboChallenge Table30-V2.}
We use human episode-level annotations from the official RoboChallenge platform.
Platform operators already rate each rollout on a 1--10 scale according to task completion, with successful trajectories typically receiving scores of at least 8.5.
We directly use these platform ratings as rollout quality scores, except that all failed trajectories are assigned $q=1$.
Expert demonstrations receive $q^\star=12$.

\textbf{Single-task shirt-folding post-training.}
We likewise use human episode-level annotation.
After each rollout, the data collector assesses task completion before resetting the scene and assigns a score on a 1--10 scale, following the same completion-based scoring approach as RoboChallenge.
Failed trajectories receive $q=1$, and task-specific expert demonstrations receive $q^\star=12$.

\textbf{RoboTwin~2.0.}
We apply coarse-level annotation separately to rollouts from expert-trained LoopWAM, X-VLA, and $\pi_{0.5}$.
For each policy, we compute its overall rollout success rate across all tasks and manually inspect approximately 0.7\% of its trajectories for motion smoothness and action accuracy.
These two assessments determine a single shared score for all successful trajectories from that policy across tasks; every failed trajectory receives $q=1$.
Expert-trained LoopWAM achieves approximately 90\% success during rollout collection and produces relatively smooth trajectories, so its successful rollouts receive $q=4$.
X-VLA and $\pi_{0.5}$ each achieve less than 50\% success and produce less smooth trajectories, so their successful rollouts receive $q=3$.
Expert demonstrations receive $q^\star=5$.

\textbf{RoboCasa365.}
We use the same coarse-level protocol, assessing each source policy through its overall rollout success rate across tasks and manual inspection of approximately 0.7\% of its trajectories for motion smoothness and action accuracy.
We obtain two expert-trained LoopWAM policies using 64-step and 32-step action chunks, denoted Policy~1 and Policy~2, respectively, and also collect rollouts from RLDX-1.
Following the same scoring criteria as RoboTwin, all successful rollouts from Policy~1 share $q=3$, those from Policy~2 share $q=4$, and those from RLDX-1 share $q=2$.
All failed rollouts receive $q=1$, and expert demonstrations receive $q^\star=5$.

\subsection{Training Details}

\textbf{Temporal alignment and robot representations.}
The training configurations use an action horizon of $H=64$ and a visual stride of eight control steps, yielding nine RGB frames including the initial observation.
Camera views are stitched into a single image, and robot states and actions are normalized using expert-data statistics.
For Table30-V2, joint targets are represented relative to the joint state at the beginning of each action chunk, whereas gripper commands remain absolute.
Only the joint components are converted back to absolute targets at execution; the gripper components are not differenced.
To reduce redundant stationary data, the real-world pipeline checks unnormalized state sequences with an absolute tolerance of $10^{-3}$ and resamples nearly static windows up to three times.
Windows starting within the final 20 seconds of expert episodes are exempt from this filtering to retain terminal behavior.
Padding is excluded from the denoising losses.

\textbf{Noise schedules and loss weighting.}
The base noise sampler uses $\phi_5(u)=5u/(1+4u)$ for $u\sim\mathcal U(0,1)$.
Samples drawn from the Beta distributions in Equation~\ref{eq:source_noise} are clipped to $[10^{-4},1-10^{-4}]$ without a further time shift.
Both modalities use the same scheduler weighting function in Equation~\ref{eq:modality_loss}:
\begin{equation}
\begin{aligned}
    g(\sigma)&=\exp\!\left[-2(\sigma-\tfrac12)^2\right]-e^{-1/2},\\
    \omega(\sigma)&=\frac{g(\sigma)}{C+10^{-10}},
    \qquad C=\frac{1}{T}\sum_{j=0}^{T-1}g\!\left(\phi_5(1-j/T)\right),
    \quad T=1000.
\end{aligned}
\end{equation}
This normalization is shared by expert and rollout samples.
For $K$ inference steps, the schedule is $\sigma_k=\phi_5(1-k/K)$, $k=0,\ldots,K$.
Each modality follows the Euler update
$X_m^{k+1}=X_m^k+(\sigma_{k+1}-\sigma_k)v_\theta^m$,
with the vector fields evaluated jointly and the initial observation latent fixed throughout sampling.

\subsection{Table30-V2 Subtask Segmentation}
\label{app:table30_training}


We use Qwen2.5-VL-7B for subtask segmentation, switching between stage-specific language instructions during online robot execution.
Table~\ref{tab:table30_subtasks} lists the resulting ordered prompts.
The four illustrated tasks are button pressing, plant watering, paper shredding, and stamp positioning. The remaining tasks retain their original full-task instructions.

\begin{table*}[t]
\centering
\caption{Table30-V2 subtask instructions, listed in execution order.}
\label{tab:table30_subtasks}
\small
\renewcommand{\arraystretch}{1.15}
\begin{tabularx}{\textwidth}{@{}llX@{}}
\toprule
Task & Robot & Ordered subtask instructions \\
\midrule
Press the button & ARX5 & 1. Press the pink button.\newline
2. Press the blue button.\newline
3. Press the green button.\newline
4. Press the yellow button, then retract the gripper. \\
\midrule
Water the flowers & ARX5 & 1. Pick up the watering can and water the potted plant.\newline
2. The watering is finished, put the watering can down. \\
\midrule
Shred paper & UR5 & 1. Put the paper into the shredder.\newline
2. Retract gripper. \\
\midrule
Stamp positioning & ALOHA & 1. Pick up the stamp.\newline
2. Dip into the ink.\newline
3. Stamp the signature area and return. \\
\bottomrule
\end{tabularx}
\end{table*}

\subsection{Subtask Switching Examples}
\label{app:subtask_examples}
\begin{figure}[tp]
\centering
\includegraphics[width=0.8\linewidth]{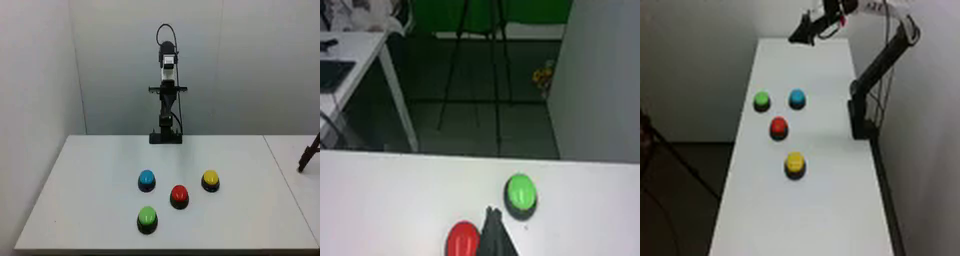}
\par\smallskip{\footnotesize\textbf{Stage 1.} Press the pink button.\par}
\medskip
\includegraphics[width=0.8\linewidth]{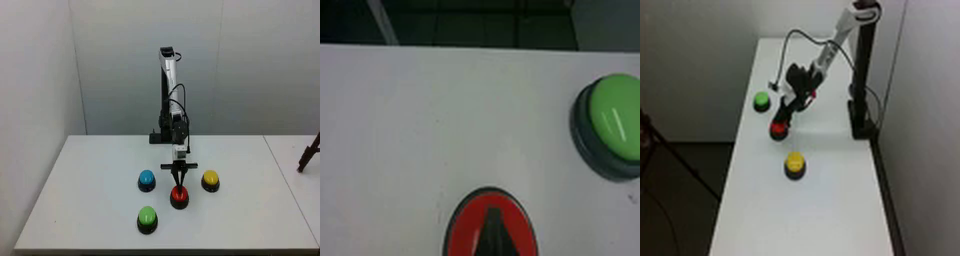}
\par\smallskip{\footnotesize\textbf{Stage 2.} Press the blue button.\par}
\medskip
\includegraphics[width=0.8\linewidth]{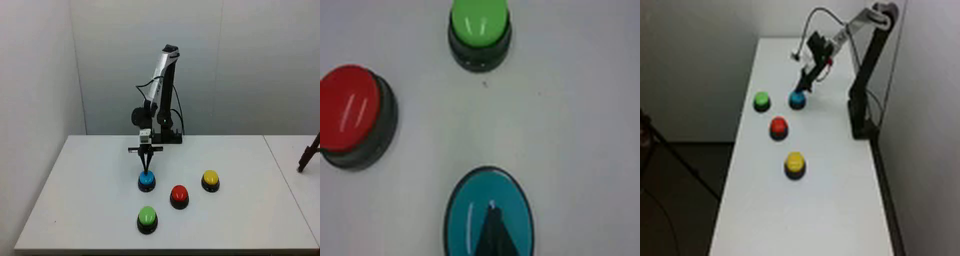}
\par\smallskip{\footnotesize\textbf{Stage 3.} Press the green button.\par}
\medskip
\includegraphics[width=0.8\linewidth]{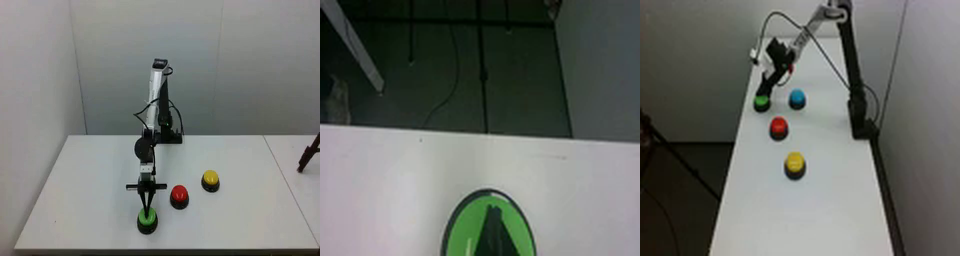}
\par\smallskip{\footnotesize\textbf{Stage 4.} Press the yellow button, then retract the gripper.\par}
\caption{\textbf{Button pressing (ARX5).} Subtask segmentation example. The initial observation (Stage 1) and subsequent stage observations are shown in order, each paired with its subtask instruction.}
\label{fig:subtask_press_the_button}
\end{figure}

\begin{figure}[tp]
\centering
\includegraphics[width=0.8\linewidth]{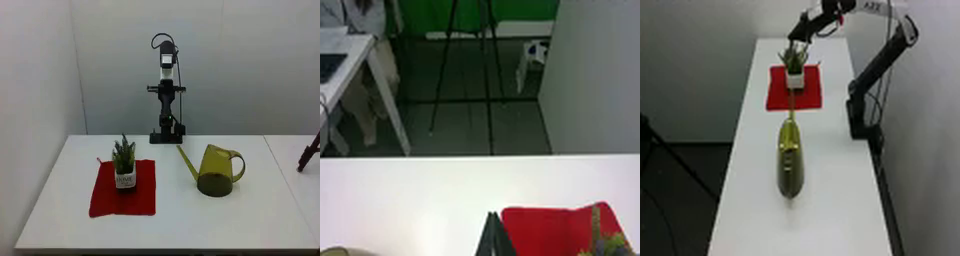}
\par\smallskip{\footnotesize\textbf{Stage 1.} Pick up the watering can and water the potted plant.\par}
\medskip
\includegraphics[width=0.8\linewidth]{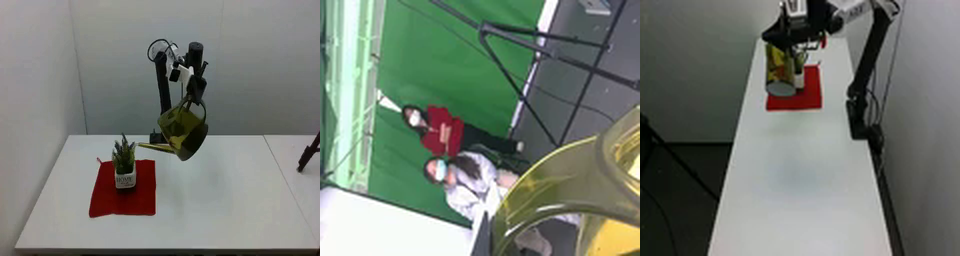}
\par\smallskip{\footnotesize\textbf{Stage 2.} The watering is finished, put the watering can down.\par}
\caption{\textbf{Plant watering (ARX5).} Subtask segmentation example. The initial observation (Stage 1) and subsequent stage observations are shown in order, each paired with its subtask instruction.}
\label{fig:subtask_water_the_flowers}
\end{figure}

\begin{figure}[tp]
\centering
\includegraphics[width=0.8\linewidth]{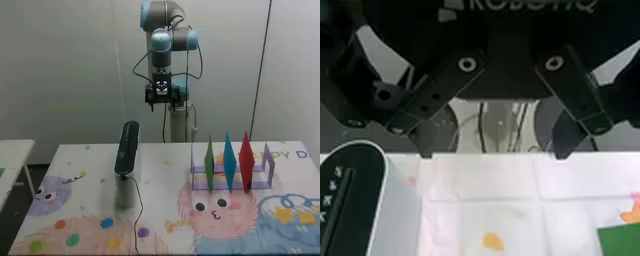}
\par\smallskip{\footnotesize\textbf{Stage 1.} Put the paper into the shredder.\par}
\medskip
\includegraphics[width=0.8\linewidth]{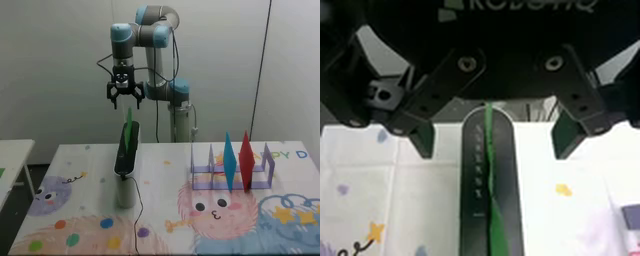}
\par\smallskip{\footnotesize\textbf{Stage 2.} Retract gripper.\par}
\caption{\textbf{Paper shredding (UR5).} Subtask segmentation example. The initial observation (Stage 1) and subsequent stage observations are shown in order, each paired with its subtask instruction.}
\label{fig:subtask_shred_paper}
\end{figure}

\begin{figure}[tp]
\centering
\includegraphics[width=0.8\linewidth]{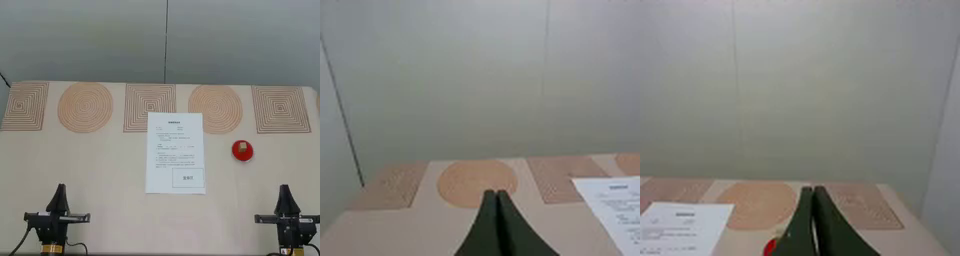}
\par\smallskip{\footnotesize\textbf{Stage 1.} Pick up the stamp.\par}
\medskip
\includegraphics[width=0.8\linewidth]{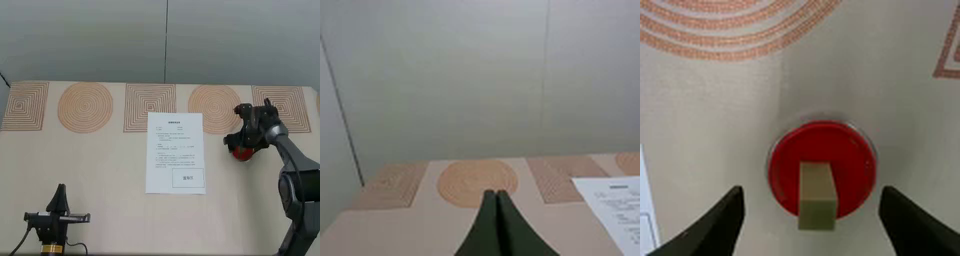}
\par\smallskip{\footnotesize\textbf{Stage 2.} Dip into the ink.\par}
\medskip
\includegraphics[width=0.8\linewidth]{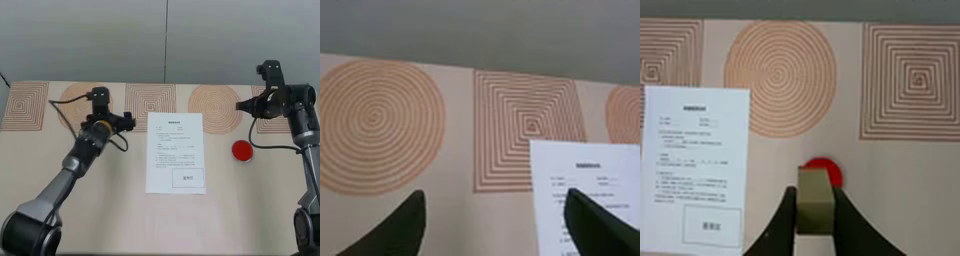}
\par\smallskip{\footnotesize\textbf{Stage 3.} Stamp the signature area and return.\par}
\caption{\textbf{Stamp positioning (ALOHA).} Subtask segmentation example. The initial observation (Stage 1) and subsequent stage observations are shown in order, each paired with its subtask instruction.}
\label{fig:subtask_stamp_positioning}
\end{figure}

Figures~\ref{fig:subtask_press_the_button}--\ref{fig:subtask_stamp_positioning} show one trajectory for each of the four tasks.
The first image is the stitched policy input at the start of execution; each subsequent image is captured when the active subtask prompt changes.
The associated text gives the instruction used for the new stage.

\section{RoboChallenge Table30-V2 Leaderboard}
\label{app:leaderboard}
\begin{figure}[t]
\centering
\includegraphics[width=0.82\linewidth]{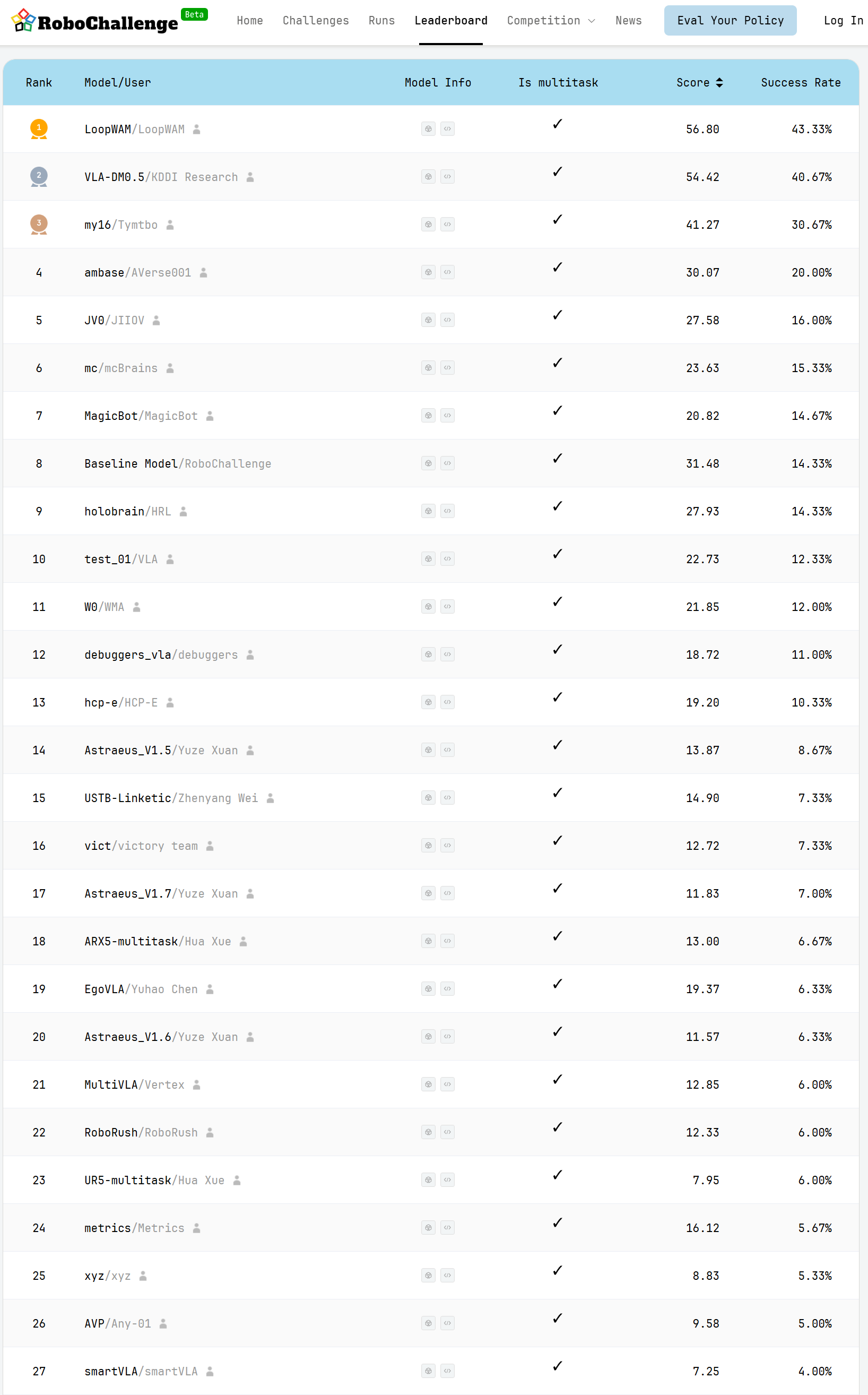}
\caption{\textbf{RoboChallenge Table30-V2 leaderboard.} LoopWAM ranks first with an overall score of 56.80 and a success rate of 43.33\%, using a multi-task policy for each embodiment.}
\label{fig:table30_leaderboard}
\end{figure}

Figure~\ref{fig:table30_leaderboard} presents the official RoboChallenge Table30-V2 leaderboard at the time of evaluation.
LoopWAM ranks first, achieving an overall score of 56.80 and a success rate of 43.33\%.

\section{Failure Recovery and Iterative Execution}
\label{app:retry}

Real-world robot manipulation often involves uncertainty caused by
imperfect perception, inaccurate execution, and complex object
interaction dynamics.

\begin{figure}[t]
    \centering

    \includegraphics[width=\linewidth]{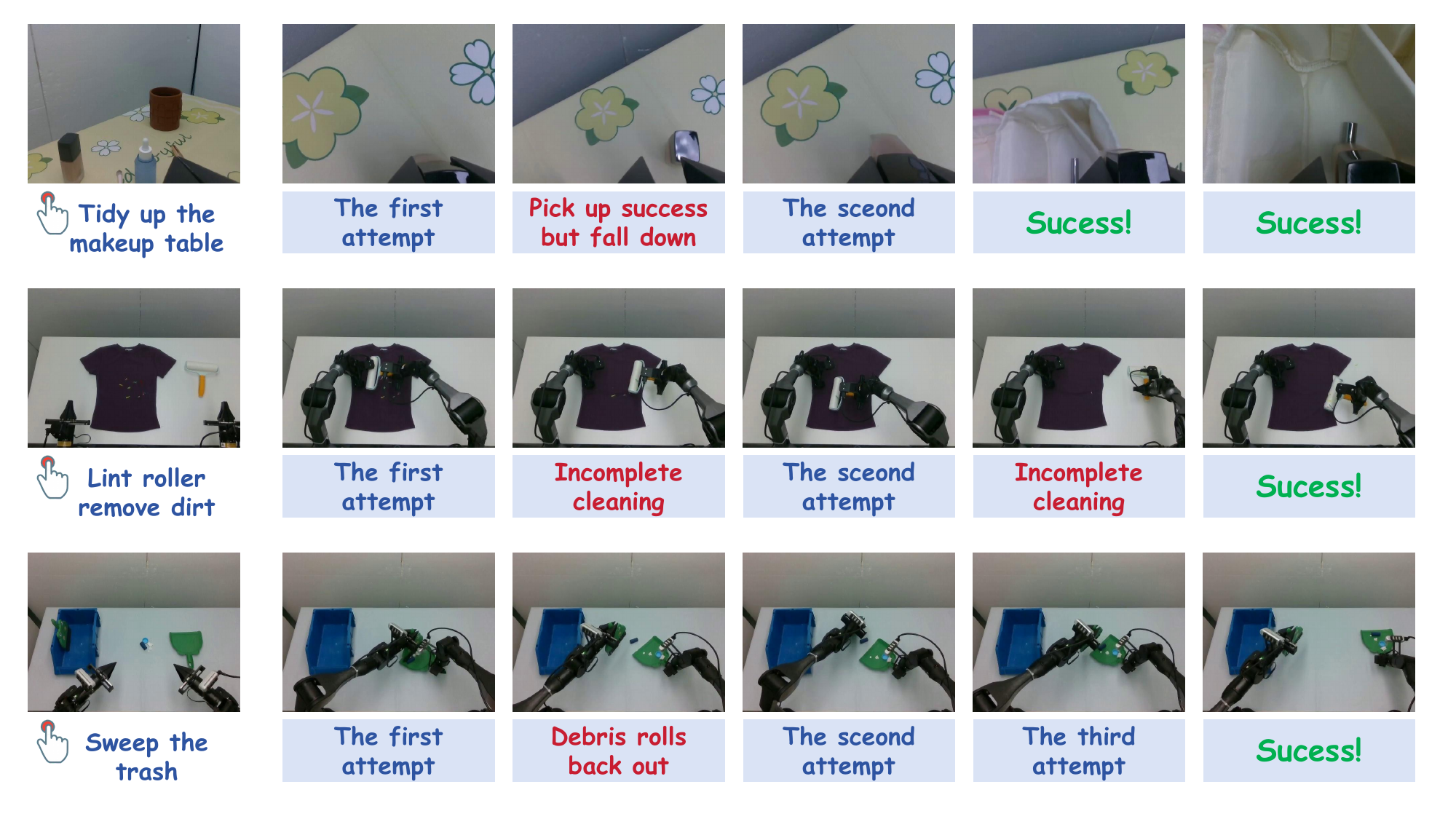}

    \caption{
        Qualitative examples of iterative execution and failure recovery.
        The first attempt may fail due to execution uncertainty,
        incomplete interaction, or object displacement.
        Subsequent attempts generate corrective behaviors and achieve
        successful task completion.
    }
    \label{fig:retry_success}
\end{figure}

To analyze the ability of LoopWAM to support iterative execution,
Figure~\ref{fig:retry_success} presents qualitative examples where an
initial attempt does not complete the desired task.

After observing the resulting environment state, the model generates
subsequent corrective behaviors that improve the execution outcome.
These examples highlight the importance of modeling future states and
environment transitions beyond predicting isolated actions.

\section{Qualitative Future Observation Prediction}
\label{app:video_prediction}

As a learned robot world model, LoopWAM predicts future visual
observations conditioned on the current robot state and interaction
history.

\begin{figure}[t]
    \centering

    \includegraphics[width=\linewidth]{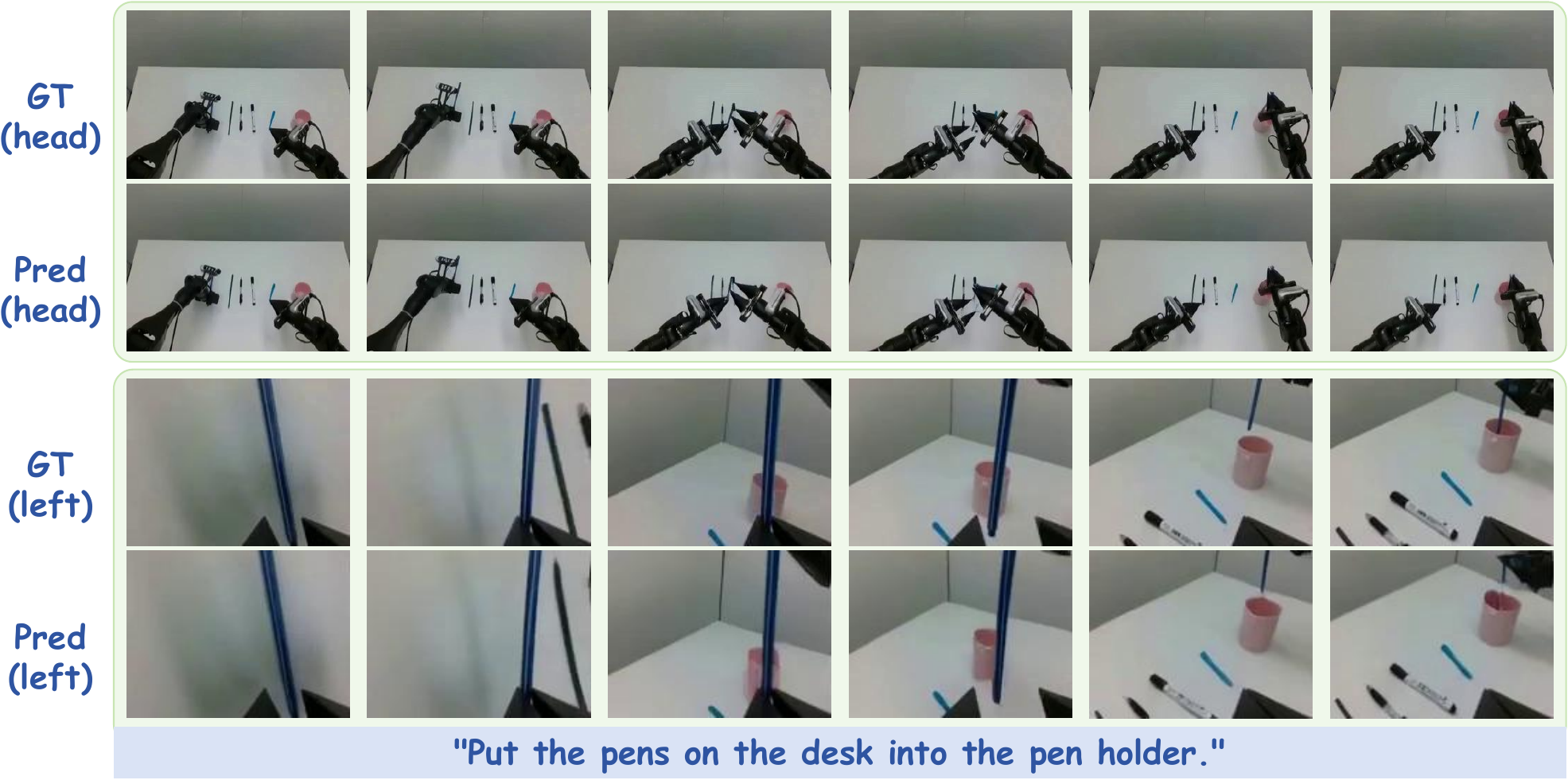}

    \vspace{0.5em}

    \includegraphics[width=\linewidth]{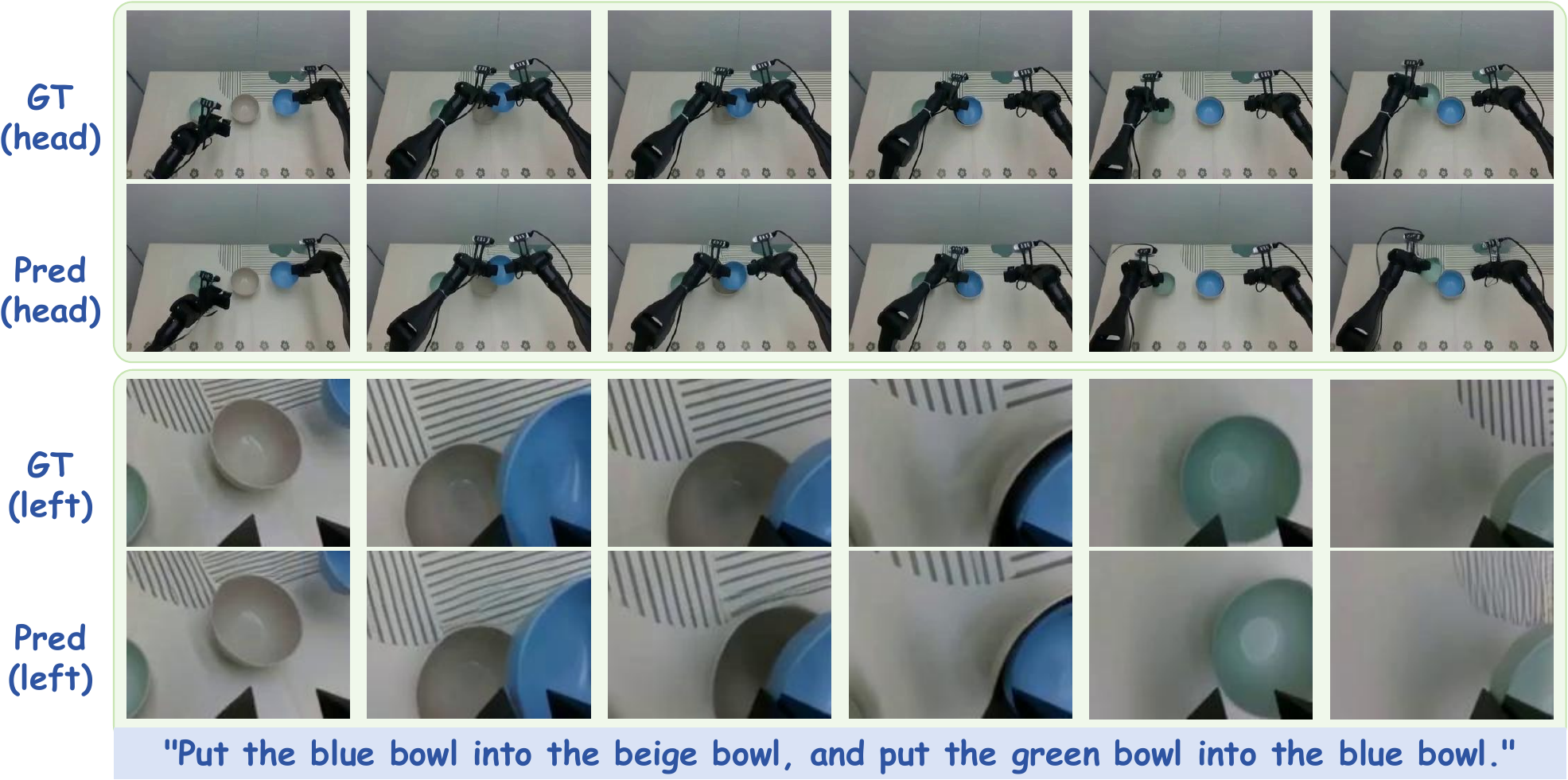}

    \caption{
        Qualitative comparison between predicted future observations from
        LoopWAM and ground-truth robot executions.
        The model captures the temporal evolution of robot-object
        interactions and predicts plausible future visual states.
    }
    \label{fig:video_prediction}
\end{figure}

To further examine the learned visual dynamics,
Figure~\ref{fig:video_prediction} compares predicted future frames with
ground-truth observations from robot executions.

The predictions capture major aspects of scene evolution, including
robot motion, object movement, and manipulation progress.
Although generated observations may differ from the ground-truth images
in fine visual details, they preserve the temporal structure of the
interaction process.

\section{Representative Robot Platforms and Tasks}
\label{app:robot_tasks}

LoopWAM is evaluated across multiple robot platforms with different
embodiments, sensing configurations, and manipulation spaces.

To demonstrate the diversity of learned robot behaviors, we provide
additional qualitative examples from ARX, ALOHA, UR5, and W1 platforms.
The examples cover object manipulation, coordinated control,
fine-grained interaction, and long-horizon tasks.

\begin{figure}[t]
    \centering

    \includegraphics[width=0.95\linewidth]{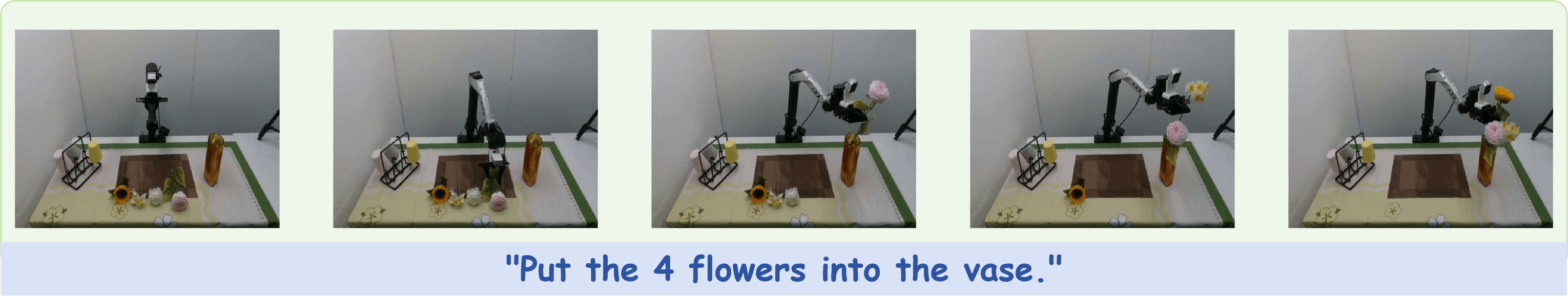}

    \vspace{0.4em}

    \includegraphics[width=0.95\linewidth]{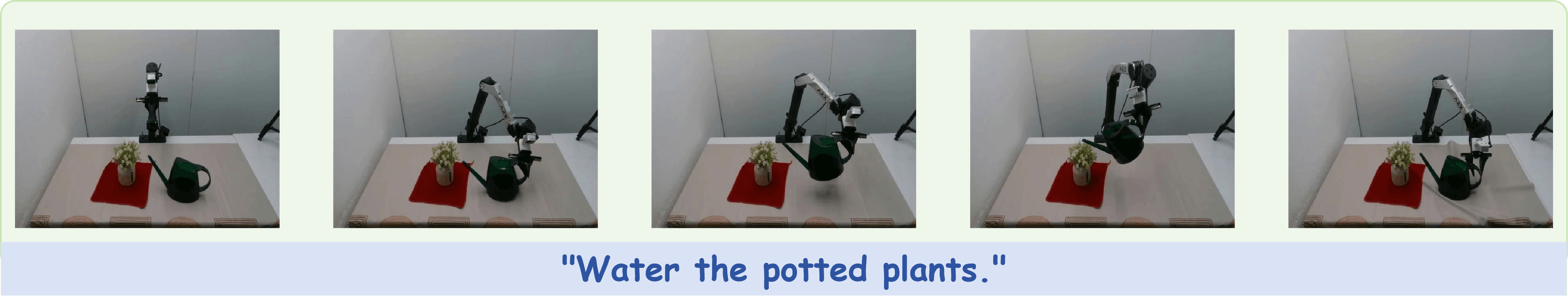}

    \caption{
        Representative tasks on the ARX robot platform.
        The examples illustrate diverse manipulation scenarios used for
        evaluating LoopWAM across different robot embodiments.
    }
    \label{fig:arx_tasks}
\end{figure}

\begin{figure}[t]
    \centering

    \includegraphics[width=0.95\linewidth]{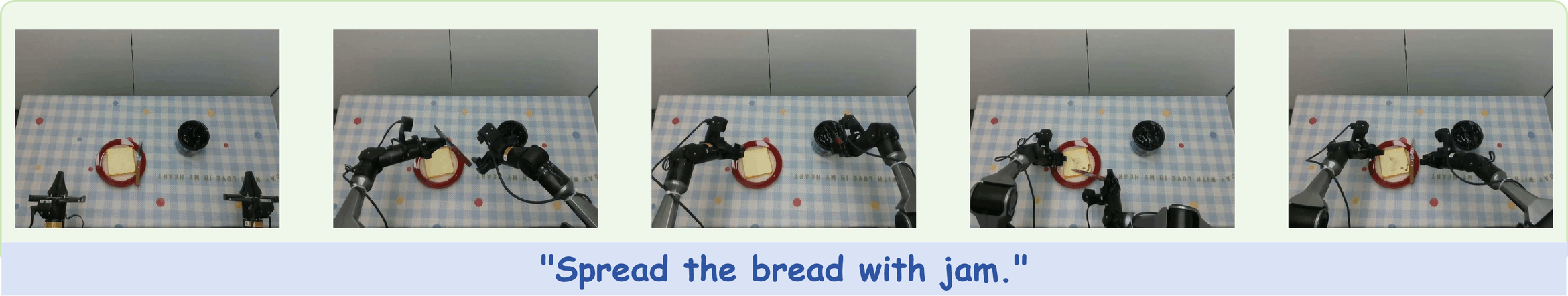}

    \vspace{0.4em}

    \includegraphics[width=0.95\linewidth]{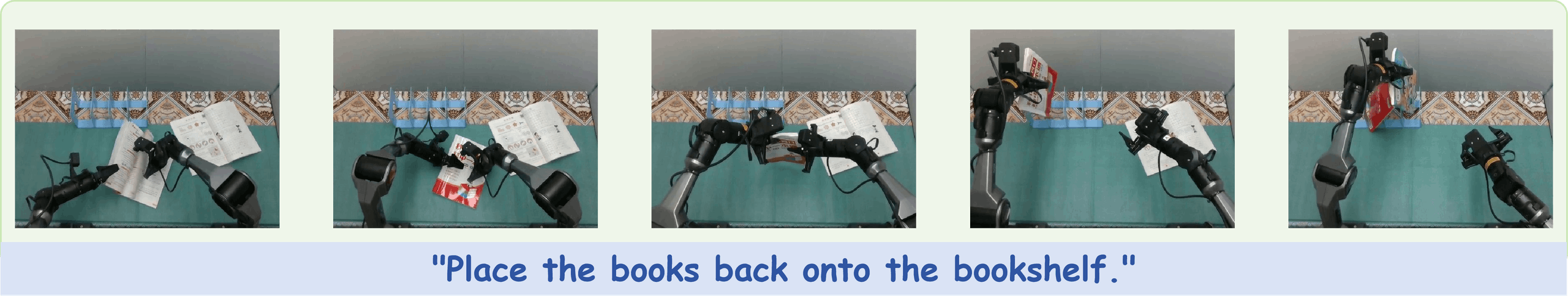}

    \vspace{0.4em}

    \includegraphics[width=0.95\linewidth]{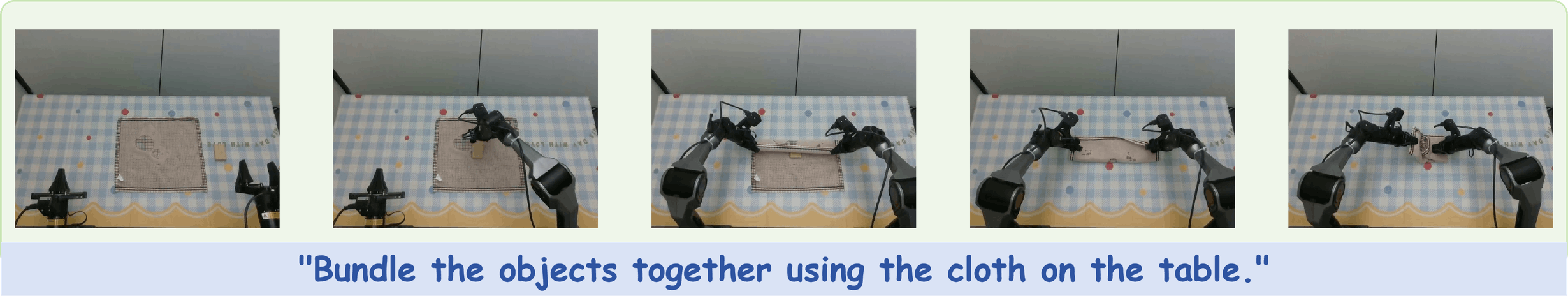}

    \vspace{0.4em}

    \includegraphics[width=0.95\linewidth]{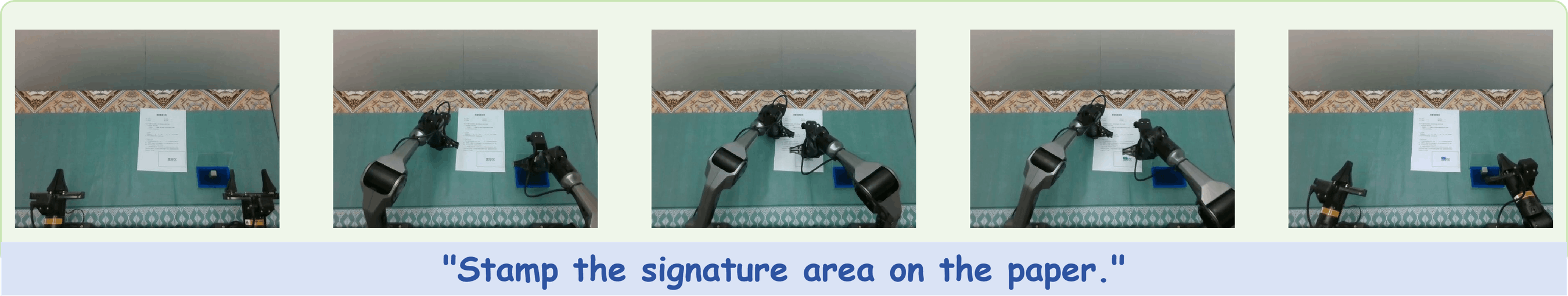}

    \caption{
        Representative tasks on the ALOHA robot platform.
        The examples include coordinated manipulation and object
        interaction requiring temporally consistent control.
    }
    \label{fig:aloha_tasks}
\end{figure}

\begin{figure}[t]
    \centering

    \includegraphics[width=0.95\linewidth]{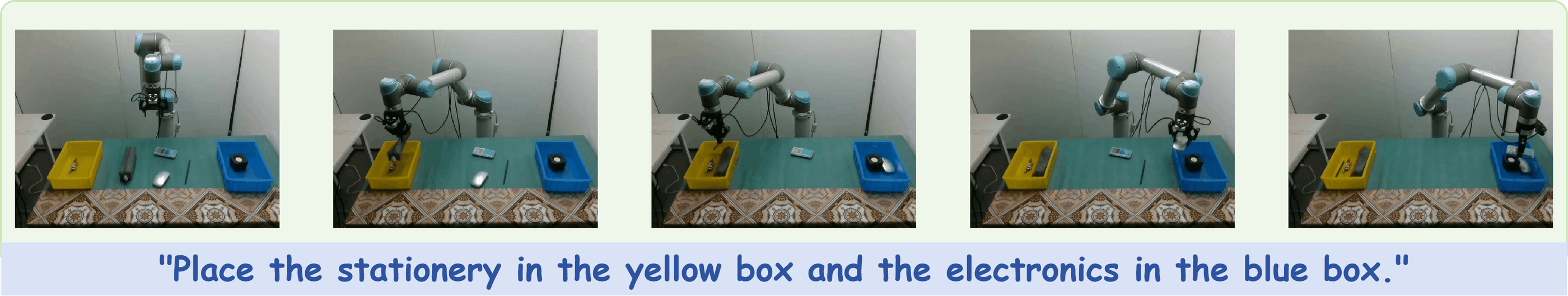}

    \vspace{0.4em}

    \includegraphics[width=0.95\linewidth]{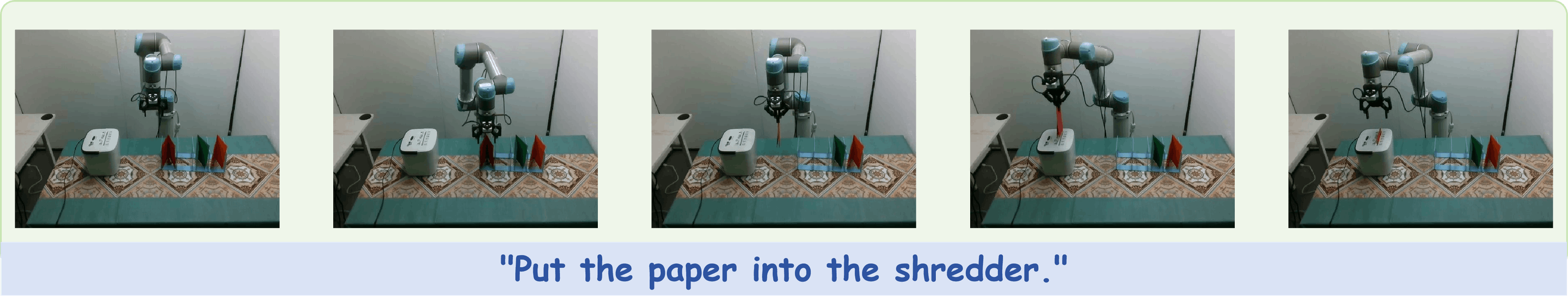}

    \caption{
        Representative tasks on the UR5 robot platform.
        The examples demonstrate single-arm manipulation in structured
        environments.
    }
    \label{fig:ur5_tasks}
\end{figure}

\begin{figure}[t]
    \centering

    \includegraphics[width=0.95\linewidth]{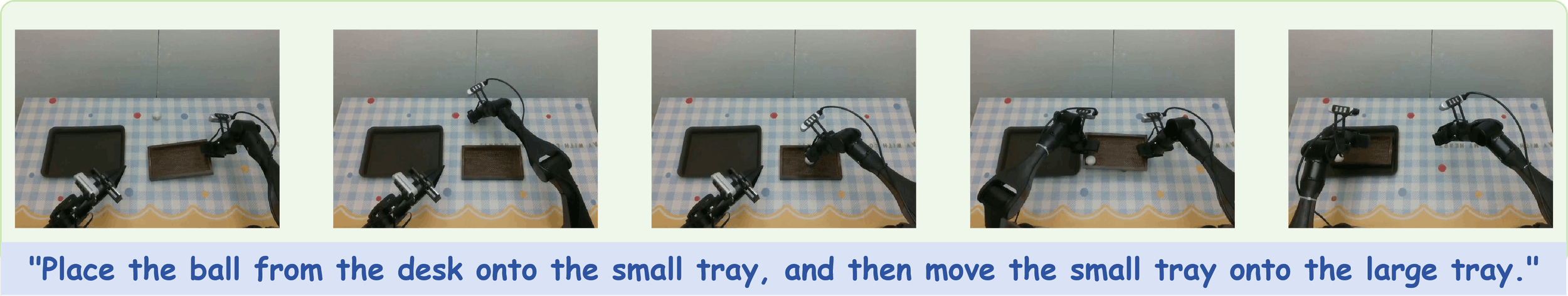}

    \vspace{0.4em}

    \includegraphics[width=0.95\linewidth]{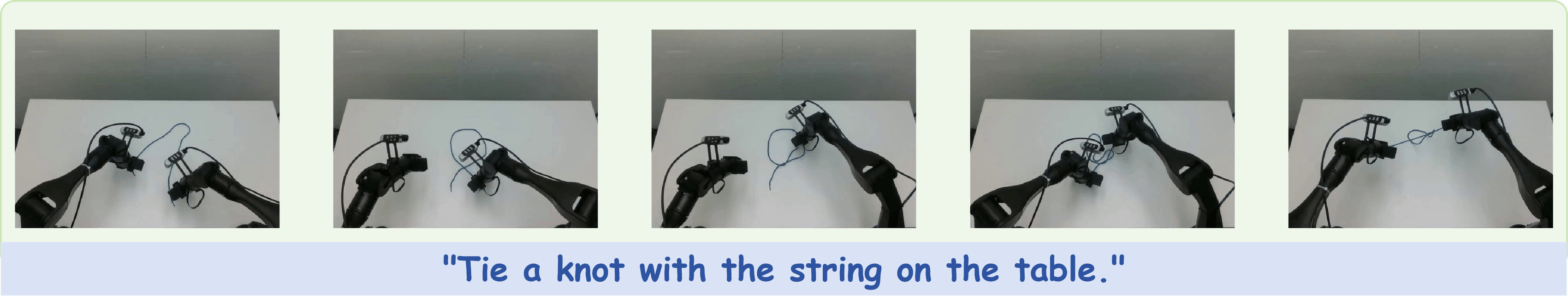}

    \vspace{0.4em}

    \includegraphics[width=0.95\linewidth]{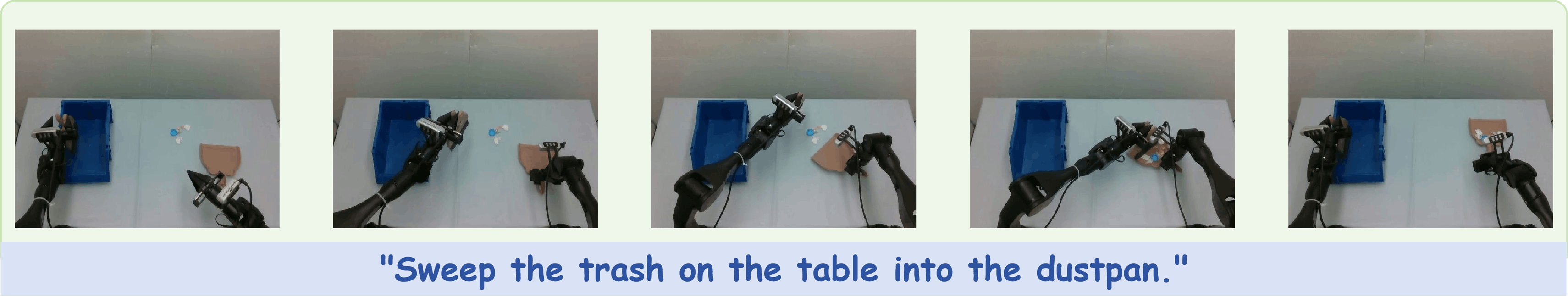}

    \vspace{0.4em}

    \includegraphics[width=0.95\linewidth]{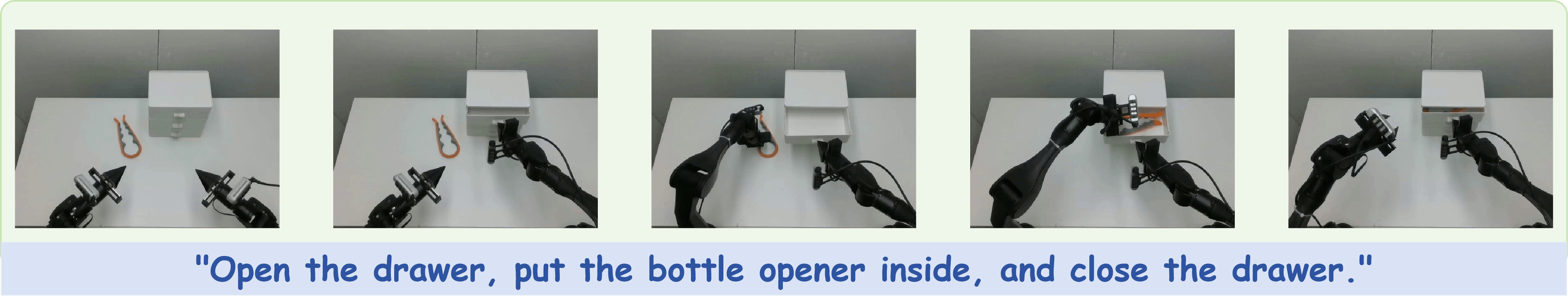}

    \caption{
        Representative long-horizon tasks on the W1 robot platform.
        The examples demonstrate sequential manipulation behaviors in
        complex environments.
    }
    \label{fig:w1_tasks}
\end{figure}

Figure~\ref{fig:arx_tasks} shows representative tasks on the ARX platform.
Figure~\ref{fig:aloha_tasks} presents examples from the ALOHA platform,
which includes coordinated manipulation behaviors.
Figure~\ref{fig:ur5_tasks} demonstrates single-arm manipulation tasks on
the UR5 platform.
Finally, Figure~\ref{fig:w1_tasks} presents representative long-horizon
tasks from the W1 platform.


\end{document}